\documentclass[10pt,twocolumn,letterpaper]{article}
\pdfoutput=1
\usepackage{cvpr}
\usepackage{times}
\usepackage{epsfig}
\usepackage{graphicx}
\usepackage{amsmath}
\usepackage{amssymb}

\usepackage{epstopdf}
\usepackage{algorithm}
\usepackage{algorithmic}
\usepackage{caption}
\usepackage{subcaption}
\usepackage{color}
\usepackage{hyperref}

\cvprfinalcopy 

\def\cvprPaperID{1681} 

\ifcvprfinal\fi
\begin{document}

\title{Semantic Part Segmentation using Compositional Model combining Shape and Appearance}

\author{Jianyu Wang, Alan Yuille\\
University of California, Los Angeles\\
{\tt\small wjyouch@ucla.edu, yuille@stat.ucla.edu}
}

\maketitle

\begin{abstract}
In this paper, we study the problem of semantic part segmentation for animals. This is more challenging than standard object detection, object segmentation and pose estimation tasks because semantic parts of animals often have similar appearance and highly varying shapes. To tackle these challenges, we build a mixture of compositional models to represent the object boundary and the boundaries of semantic parts. And we incorporate edge, appearance, and semantic part cues into the compositional model. Given part-level segmentation annotation, we develop a novel algorithm to learn a mixture of compositional models under various poses and viewpoints for certain animal classes. Furthermore, a linear complexity algorithm is offered for efficient inference of the compositional model using dynamic programming. We evaluate our method for horse and cow using a newly annotated dataset on Pascal VOC 2010 which has pixelwise part labels. Experimental results demonstrate the effectiveness of our method.
\end{abstract}

\section{Introduction}
The past few years have witnessed significant progress on various object-level visual recognition tasks, such as object detection \cite{felzenswalb10, fidler13}, object segmentation \cite{Smin12cls, arbelaez2012semantic}, etc. Understanding how different parts of an object are related and where the parts are located have been an increasingly important topic in computer vision. There is extensive study on some part-level visual recognition tasks, such as human pose estimation (predicting joints) \cite{yang2011articulated, toshev2013deeppose} and landmark localization (predicting keypoints) \cite{belhumeur2011localizing, liubird}. But there are only a few pieces of works on semantic part segmentation, such as human parsing \cite{bo2011shape, dongdeformable, dongtowards, yamaguchi2012parsing} and car parsing \cite{thomas2008using, eslami2012generative, Lu2014}. In some applications (e.g., activity analysis), it would be of great use if computers can output richer part segmentation instead of just giving a set of keypoints/landmarks, a bounding box or an entire object segment. 
\begin{figure}[t]
		\begin{center}
    		\includegraphics[width=0.4\textwidth]{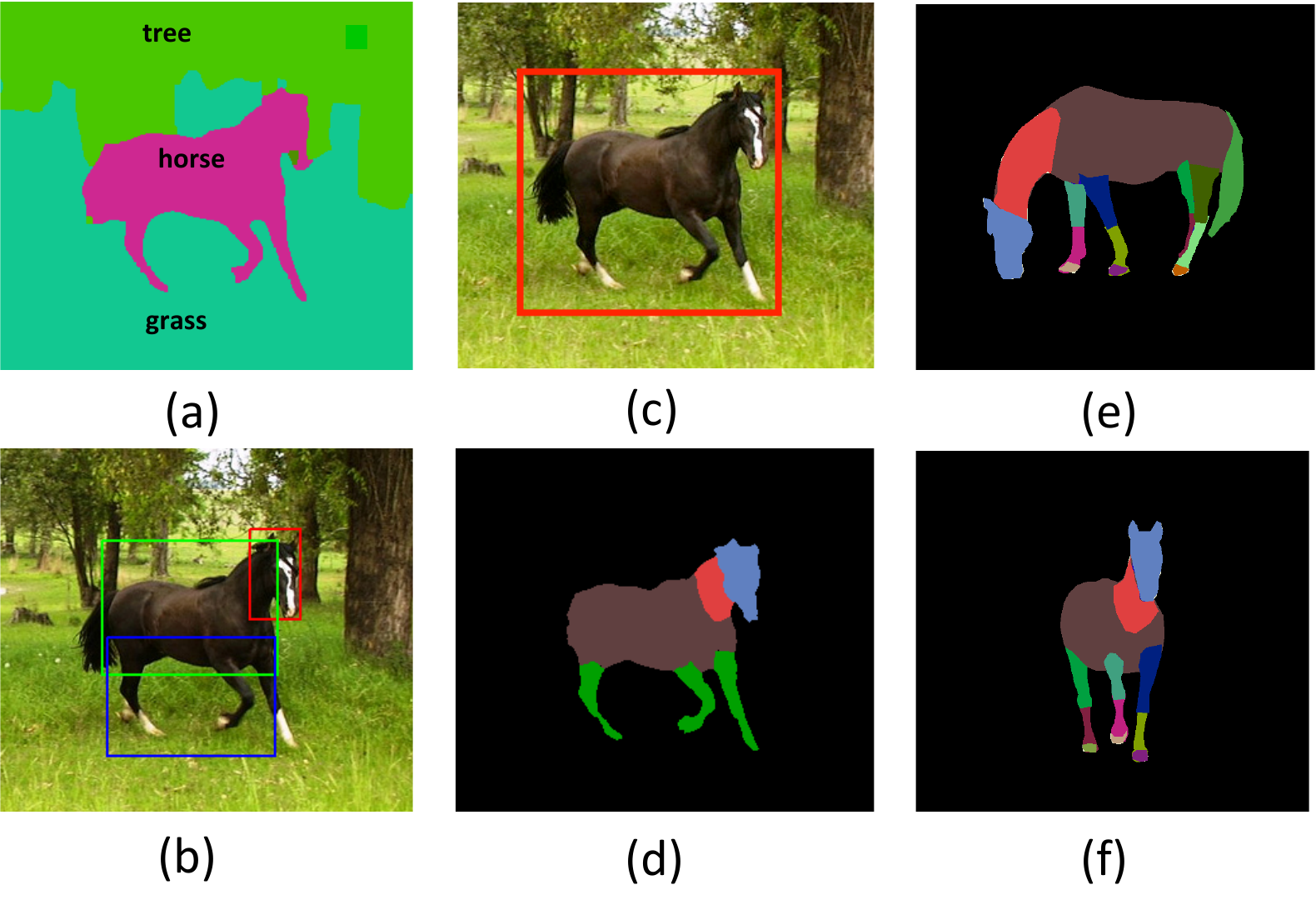}
    		\caption{Different visual recognition tasks: (a) semantic labeling with pixelwise object and background label. (b) object detection which outputs bounding box. (c) part detection which gives bounding box for part. (d,e,f) semantic part segmentation with pixelwise part label. We study the semantic part segmentation problem in this paper. Best viewed in color.} \label{fig:introfigure}
		\end{center}
\end{figure}

We make an attempt on the challenging task of semantic part segmentation for animals in this paper. Since animals often have homogeneous appearance (e.g., furs) on the whole body, mid-level segmentation methods \cite{Carreira12, Arbelaez11} could not output quality proposals for semantic parts. Besides, current classifers are not able to distinguish between different semantic parts since they often have similar appearance. Thus we could not simply take the popular object segmentation pipeline \cite{Smin12cls, arbelaez2012semantic} by treating each semantic part as an object. This tells us that the shape information of semantic parts are necessary for part segmentation. But there is a large amount of variability of part shapes due to different animal viewpoints and poses (see (d,e,f) in Figure \ref{fig:introfigure}). Therefore, it is very challenging to build a model that effectively combines animal appearance, part shape and spatial relation among parts under varying viewpoints and poses, while still allowing efficient learning and inference.  

Inspired by \cite{jin2006context,zhu2007stochastic,zhu2010learning,kokkinos2011inference}, compositional model is able to capture long-range relations among parts while still enabling efficient inference since parts are arranged in a hierarchical manner. The intuition of compositional model is that articulated objects are often built by compositions of the parts, which in turn are built by compositions of more elementary subparts. Specifically, in this paper, we build a mixture of compositional models  to represent the animal and part shapes/boundaries.  Each mixture is able to handle local deformation of shapes and different mixtures deal with global variations due to viewpoints and poses.  We incorporate edge, appearance, and part cues into the compositional model by using algorithms from edge detection, semantic labeling and part detection. 

It is of significant importance to design efficient \textit{inference} and \textit{learning} algorithms for the compositional model. We develop the constrained generalized distance transform (CGDT) algorithm which extends the distance transform algorithm in \cite{felzenszwalb2004distance}. This algorithm allows us to perform efficient linear-time inference for the model. Besides, we design a novel algorithm to learn the compositional models for animal and parts boundaries under various poses and viewpoints from the part-level annotation. And we learn the parameters of the model using latent SVM. 

In order to segment highly deformable animal legs, we first perform part segmentation using our compositional model for large parts, such as head, neck and torso, etc. Given these segmentation results, we can narrow down the search region for legs since legs are almost always underneath the torso. Then we segment legs by combining symmetric structure and appearance information.   

Our experiment is conducted on two animal classes: horse and cow. We use a newly annotated dataset on Pascal VOC 2010 \cite{chendetect} which provides pixelwise semantic part annotations. We focus on segmenting fully observable animals in this paper and leave the occlusion and truncation issue for future study. Self-occlusion due to poses/viewpoints can be handled by our model. We compare our algorithm with the method that combines the state-of-the-art animal part detection \cite{chendetect} and object segmentation \cite{BharathECCV2014}. The experiment shows that our method achieves much better part segmentation than the baseline, which demonstrates the effectiveness of our method. 

In summary, our contribution is threefold. Firstly, we develop a novel method for animal part segmentation by introducing a mixture of compositional models coupled with shape and appearance. Secondly, we propose an algorithm to learn the compositional models of object and part shapes given part-level pixelwise annotations. Thirdly, we develop the constrained generalized distance transform (CGDT) algorithm to achieve linear-time inference for our model.


\section{Related Work}
In terms of method, our work is related to \cite{zhu2010learning, zhu2011max}, where they used compositional model for horse segmentation. But they did not incorporate strong appearance cues into their compositional shape model, and they modeled only a few poses and viewpoints. Besides, our inference is much faster than their compositional inference algorithm. There was also work on automatically learning the compositional structure/hierarchical dictionary \cite{zhu2008unsupervised, fidler2007towards}, but their algorithms did not consider semantic parts and were not evaluated on challenging datasets.

In terms of task, our work is related to human parsing/clothes parsing \cite{bo2011shape, dongdeformable, dongtowards, yamaguchi2012parsing}. They generated segment proposals by superpixel/over-segmentation algorithms, and then used these segments as building blocks for whole human body by either compositional method or And-Or graph. Note that our task is inherently quite different from clothes parsing because animals often have roughly homogeneous appearance throughout the whole body while in the human parsing datasets humans often have different appearance (e.g., color) for different part due to clothes. So their superpixel/over-segmentation algorithms could not output good segment proposals for animal parts. Besides, in challenging datasets like Pascal VOC, cluttered background and unclear boundaries further degrade the superpixel quality. Therefore, the superpixel-based methods for human parsing are not appropriate for our animal part segmentation task. 

Our work also bears a similarity to \cite{zhu2012face} in the spirit that a mixture of graphical models are used to capture global variation due to viewpoints/poses. But our compositional model is able to capture spatial relation between children nodes while still achieving linear complexity inference, and we develop an algorithm to learn the mixtures of compositional models. Besides, our task is part segmentation for animals of various poses and viewpoints, which appears more challenging than landmark localization for faces in \cite{zhu2012face}.  

There are lots of works in the literature on modeling object shape such as \cite{shotton2005contour, ferrari2010images, wu2010learning, kokkinos2007unsupervised}. But they were only aimed at object-level detection or segmentation. Furthermore, none of them combined shape representation with strong appearance information.

\section{Compositional Model combining Shape and Appearance}
We develop a compositional model to represent animal shape/boundary under various viewpoints and poses. We formulate the compositional part-subpart relation by a probabilistic graphical model. Let $v$ denote the parent node which represents the part and $ch(v)$ denote the children nodes which represent the constituent subparts. The location of part $v$ is denoted by $S_v = (x_v, y_v)$ and the locations of its subparts $ch(v)$ are denoted by $S_{ch(v)}$.  The probability distribution for the part-subpart composition is modeled as a Gibbs distribution

\begin{equation}
P(S_{ch(v)}|S_v) = 
\begin{cases}
\frac{1}{Z}{\exp(-\psi (S_{ch(v)}))}, &\text{if} \ \  S_v=f(S_{ch(v)}) \\
0 , &\text{otherwise.}
\end{cases}
\end{equation}

Here $f(S_{ch(v)})$ is a deterministic function. In this paper, we limit the number of subparts for any part to be two, i.e., $|ch(v)|=2$. And we set $f(S_{ch(v)}) = S_{ch(v)}/2$, which means that the location of a part is the average location of its children subparts. Potential function $\psi(S_{ch(v)})$ represents the relation between two children subparts. Let $ch(v) = (v_1,v_2).$ We have
\begin{equation}
\psi(S_{ch(v)})=w_v \cdot (dx_v^2,dy_v^2 ),
\end{equation}
where $dx_v = x_{v_2} -x_{v_1} -\Delta x_v $ and $dy_v = y_{v_2} -y_{v_1} -\Delta y_v$. Here $\Delta S_v = (\Delta x_v, \Delta y_v)$ is the location of part $v$'s second subpart $v_2$ relative to its first subpart $v_1$. And $(dx_v, dy_v)$ is the location displacement of second subpart $v_2$ relative to its anchor location. 

In summary, a part node $v$ is uniquely specified by its children $ch(v)$ and the spatial relation $\Delta S_v$ between children. In terms of the parent-children relation, our compositional model is similar to the prevailing pictorial structure \cite{felzenszwalb2005pictorial} and deformable part model \cite{felzenswalb10}. But our model is able to capture mutual relation between children.


\begin{figure}[t]
\centering
\begin{subfigure}[b]{150pt}
\includegraphics[width=150pt]{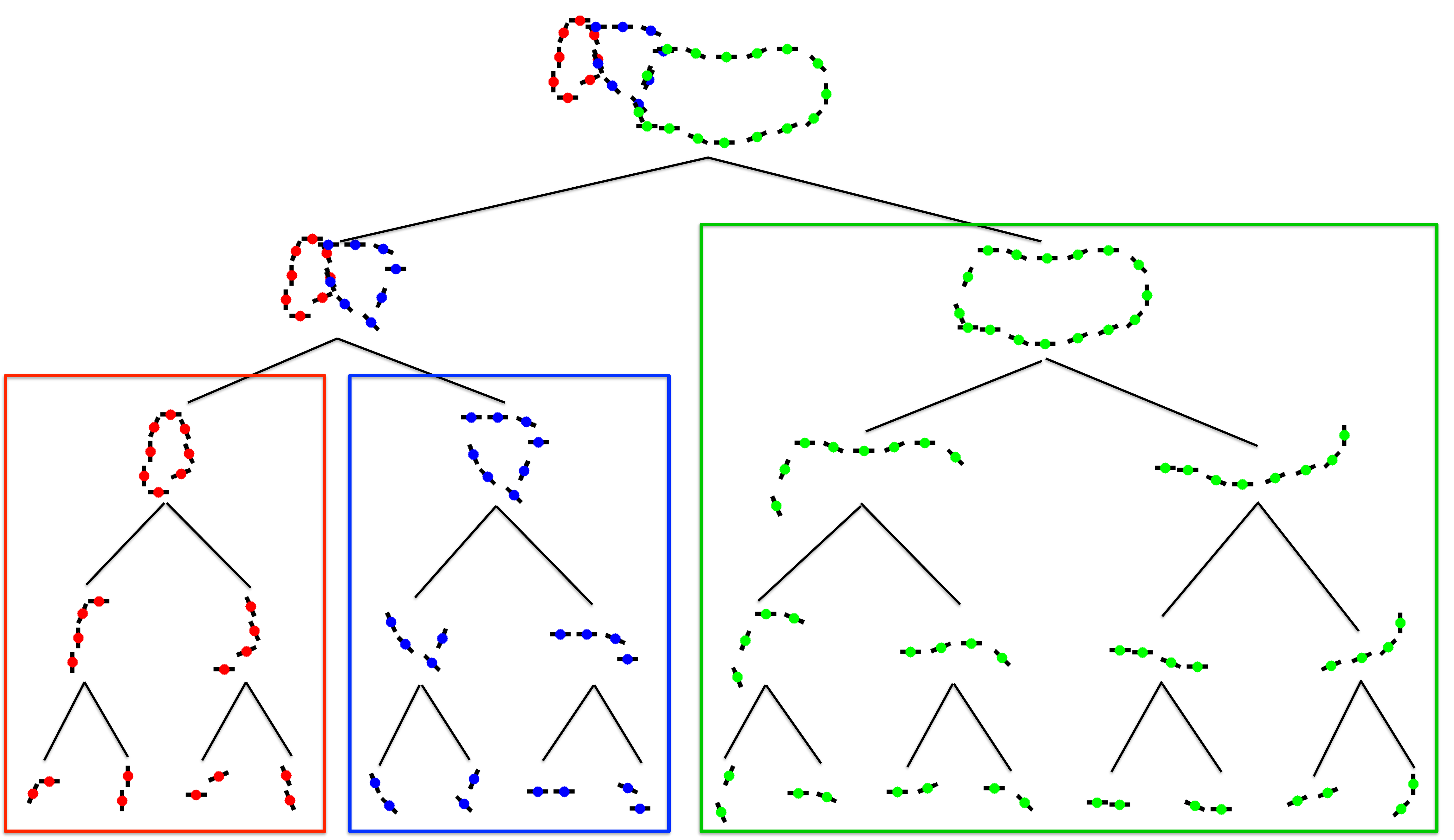}   		
    		\caption{} 
\end{subfigure}\hspace{0cm}
\begin{subfigure}[b]{80pt}
\includegraphics[width=80pt]{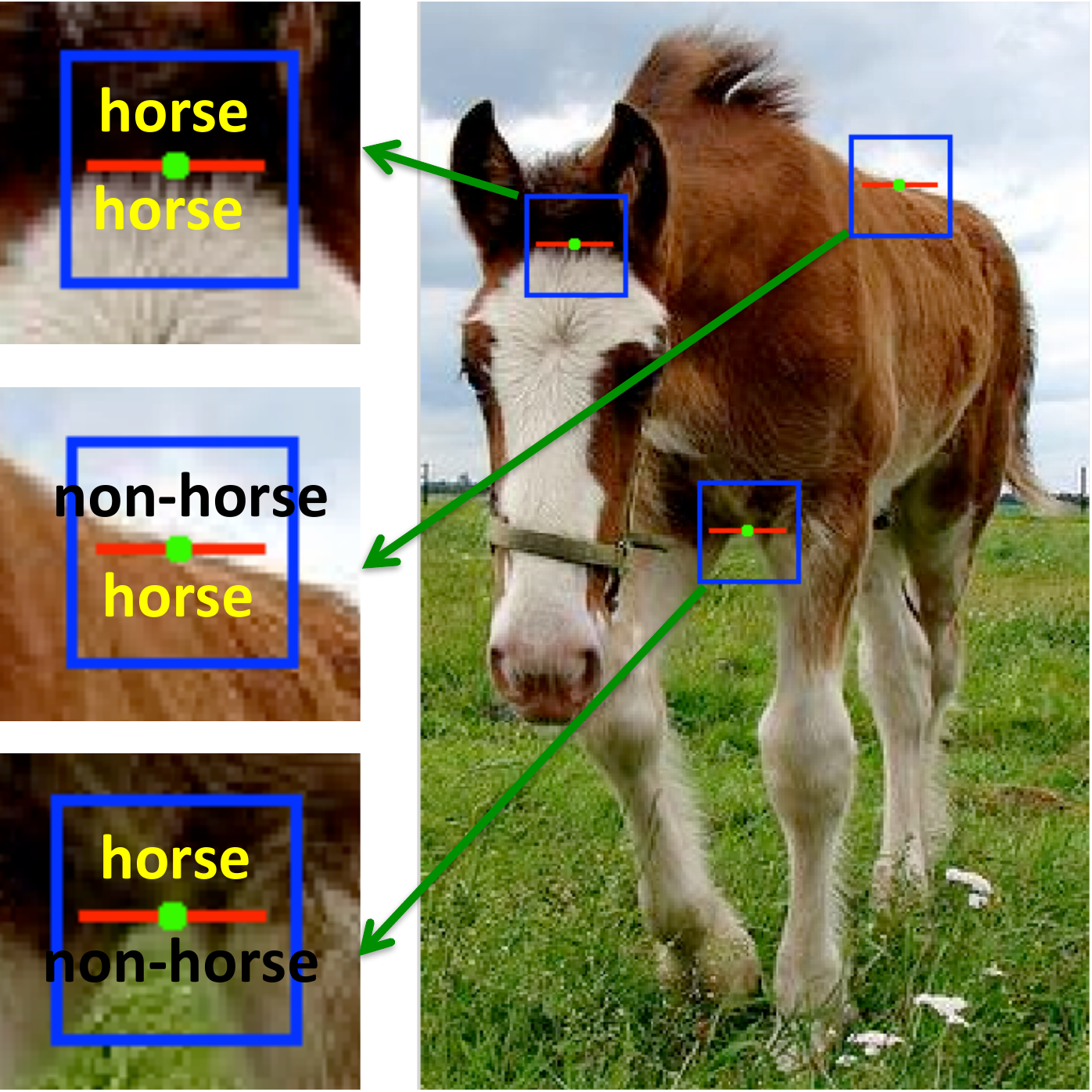}			
    		\caption{} 
\end{subfigure}  
\caption{(a) Illustration of compositional model for a particular horse shape. Red for head, blue for neck and green for torso. Due to space limitation, the leaf nodes (oriented edgelet of eight orientations) are not shown. (b) Three types of polarity value for a leaf node with a horizontal orientation. Green dot represents center location and red line segment represents orientation. Best viewed in color.}
\label{fig:modelfigure}
\end{figure}

An object can be modeled by repeating the part-subpart compositions, as shown in Figure \ref{fig:modelfigure} (a). Mathematically, we use a probabilistic graph $\mathcal{G} = (\mathcal{V},\mathcal{E})$ to model the object. This graph has a hierarchical structure with levels $l \in \{1,...,L\}$, and $\mathcal{V} = \cup_{l=1}^L {\mathcal{V}_l}$, where $\mathcal{V}_l$ denotes the node set at level-$l$. At the top level (level-$L$), there is one root node, representing the object (i.e., $|\mathcal{V}_L|=1$). The leaf node is $v \in \mathcal{V}_1$. If a part node $v$ is at level-$l$, i.e., $ v \in \mathcal{V}_l$, then its children subparts must be at level-$(l-1)$, i.e., $ch(v) \subset \mathcal{V}_{l-1}$. And as mentioned above, for any part, we limit the number of subparts to be two. So there are in total $2^{L-1}$ leaf nodes and $2^L-1$ nodes in the graph $\mathcal{V}$. Any part-subpart pair constructs an edge in this graph, i.e., $(v,t) \in \mathcal{E}$ if $t \in ch(v)$.  There is no edge between two children subparts of one parent part, i.e., $(s,t) \notin \mathcal{E}$ if $s \in ch(v), t \in ch(v)$. Thus the hierarchical graph $\mathcal{G}$ has a tree-like structure\footnote{Precisely, it is not a tree since two children are connected. But we prefer calling it tree structure in this paper for explanation purposes.}. The probability distribution for the object is specified by products of part-subpart probabilities
\begin{equation}
P(S_{\mathcal{V}}) = \prod_{v \in \mathcal{V} \setminus \mathcal{V}_1}{ P(S_{ch(v)}|S_v)} P(S_{\mathcal{V}_L}).
\end{equation} 
We assume $P(S_{\mathcal{V}_L})$ is uniformly distributed. The compositional model introduced above can be viewed as a prior shape model for the object since it characterizes the spatial relation between parts and subparts. To specify a fully generative model for the object, we need to define a likelihood function 
\begin{equation}
P(\mathbf{I} | S_{\mathcal{V}}) = \frac{1}{Z}{\exp(-\sum_{v \in \mathcal{V}}{ \phi(S_v,\mathbf{I})}))}.
\end{equation}
The MAP inference performs
\begin{equation}
\max_{S_{\mathcal{V}}} P(S_{\mathcal{V}}|\mathbf{I})  \propto P(\mathbf{I} | S_{\mathcal{V}}) P(S_{\mathcal{V}}) ,
\end{equation}
which is equivalent to minimizing the energy 
\begin{equation}
\begin{split}
E(\mathbf{I}) = \min_{S_{\mathcal{V}}} E(S_{\mathcal{V}},\mathbf{I}) 
= \sum_{v \in \mathcal{V}}{ \phi(S_v,\mathbf{I})} + \sum_{ \substack{v \in \mathcal{V} \setminus \mathcal{V}_1 \\ S_v=f(S_{ch(v)})}} {\psi(S_{ch(v)})}.   \label{eq:energymin}   
\end{split}
\end{equation}

\subsection{Feature for Unary Term}
Next we explain the potential function (unary term) which interacts with the image. We assume that the parameters are shared by parts which will be discussed in detail in Section \ref{sec:paralearning}. Specifically, the potential function for leaf node $v \in \mathcal{V}_1$ is modeled as
\begin{equation}
\phi(S_v,\mathbf{I}) = \phi^{\text{edge}}(S_v,\mathbf{I}) + \phi^{\text{app}}(S_v,\mathbf{I}).
\end{equation}
The first term $\phi^{\text{edge}}(S_v,\mathbf{I})$ characterizes how well the orientation at location $S_v$ in the image matches the model orientation $\theta_v$. In the experiment we use the gPb edge detection result \cite{Arbelaez11} which outputs pixelwise confidence score for eight orientations. Thus 
\begin{equation}
\phi^{\text{edge}}(S_v,\mathbf{I}) = w_v^{\text{edge}} \cdot gPb(\theta_v,S_v, \mathbf{I}).
\end{equation}
To incorporate appearance information, each leaf node $v$ is associated with an polarity value $a_v$ (specified by the model as $\theta_v$) indicating which side of the leaf node is object side, and which side is non-object (background) side. We extract a square centered at location $S_v$, and obtain the object-side region and non-object-side region based on the orientation, as shown in Figure \ref{fig:modelfigure} (b). We use the semantic labeling result \cite{mottaghirole} as the appearance feature. It gives pixelwise segmentation result for 34 classes including 20 object classes from Pascal VOC and another 14 background classes. Each pixel is associated with a 34-dimensional vector with each component being the confidence score for the corresponding class. We average the feature vector of object-side region and non-object-side region, and then concatenate them to make a 68-dimensional feature vector denoted by $SemLab(\theta_v,S_v, \mathbf{I})$. We use the confidence scores of all classes to deal with inaccurate semantic labeling and context information. Thus we have 
\begin{equation}
\phi^{\text{app}}(S_v,\mathbf{I}) = w_v^{\text{app}} \cdot SemLab(\theta_v,S_v, \mathbf{I}).
\end{equation}
For the non-leaf node $v \in \mathcal{V}_l, l > 1$, the unary term $\phi(S_v,\mathbf{I}) $ indicates the confidence of part $v$ being at location $S_v$. The confidence score can be from some part detection algorithm \cite{chendetect} for animals.
\begin{equation}
\phi(S_v,\mathbf{I}) = w_v^{\text{part}} \cdot PartDet(S_v, \mathbf{I}).
\end{equation} 
For example, if $v$ represents the horse head, $PartDet(S_v, \mathbf{I}) $ can be the horse head detection score.

\subsection{Mixture of Poses and Viewpoints}
We have so far introduced a compositional model for animal of one single viewpoint and pose. In order to model various poses and viewpoints, we use a set of nodes at the top level ($v \in \mathcal{V}_L$), each of which represents animal shape from one viewpoint and pose. Basically we use a mixture model with each mixture being a node at the top level. Section \ref{sec:structlearn} will introduce how to learn the mixtures. 


\section{Inference for Compositional Model}
Given an image, the goal of inference is to find the best mixture $v \in \mathcal{V}_L$ (i.e. the best viewpoint and pose) and specify locations of all its descendants $S_{tree(v)}$, especially locations of all leaf nodes as boundary landmarks. Then we can connect adjacent landmarks of each semantic part to give part segmentation result. Basically, for each mixture $v \in \mathcal{V}_L$, we solve the minimization problem (\ref{eq:energymin})  by standard dynamic programming on the tree $tree(v)$. And then we select the mixture with the minimal energy as the best mixture. The dynamic programming algorithm involves a bottom-up process starting from the leaf nodes to find the minimal energy, which is followed by a top-down process to find the best configuration. 

The search is done over every pixel in the image grid. Denote the image grid by $\mathcal{D}=\{1,...,W\} \times \{1,...,H\}$, and the size of the image grid is $|\mathcal{D}| = W \times H$. The core of dynamic programming is to solve the following minimization problem for each non-leaf node 
\begin{equation}
\begin{split}
E(S) =  &\min_{\substack{ \{ S_{1},S_{2} \} \\2 S = S_{1}+S_{2}}} { \phi(S_{1},S_{2}) + E_{1}(S_{1}) + E_{2}(S_{2}) } \\
= &\min_{\substack{ \{ S_{1} \} \\ 2S-S_{1} \in \mathcal{D} }} { \phi(S_{1},2S - S_{1}) + E_{1}(S_{1}) + E_{2}(2S - S_{1})}. \label{eq:originprob}
\end{split}
\end{equation}
Here $S$, $S_1$ and $S_2$ denote the locations of the parent (part) node and the two children (subpart) nodes respectively. $E(S)$, $E(S_1)$ and $E(S_2)$ denote the energy functions in the dynamic programming. For simplicity, we drop the subscript $v$ since it applies to every non-leaf node.  Exact solution of problem (\ref{eq:originprob}) requires quadratic complexity $O(|\mathcal{D}|^2)$, which is too slow in practice. This drives us to design an algorithm to achieve linear complexity $O(|\mathcal{D}|)$. Therefore we approximate problem (\ref{eq:originprob}) by 
$$
E(S) \approx \min_{\substack{ \{ S_{1} \} \\ 2S-S_{1} \in \mathcal{D} }}   { \phi(S_{1},2S - S_{1}) + E_{1}(S_{1}) + E_{2}(2S - S^*_{1})}, 
$$
\begin{equation}
S^*_1 = \arg \min_{\substack{ \{ S_{1} \} \\ 2S-S_{1} \in \mathcal{D} }}  {\phi(S_{1},2S -S_{1}) + E_{1}(S_{1}) }. \label{eq:newprob}
\end{equation} 
\begin{figure*}[!t]
\centering
\begin{equation}
(x_1^*,y_1^*) =  \arg \min_{\substack{ \{y_{1}\} \\1 \leq 2y-y_{1} \leq H   }}   \{ \ \  4w^y(y-y_{1}-\frac{\Delta y}{2})^2 + \min_{\substack{ \{ x_{1} \}\\ 1 \leq 2x-x_{1} \leq W }}  {4w^x(x-x_{1}-\frac{\Delta x}{2})^2 + E_{1}(x_{1},y_{1}) } \  \ \}. \label{eq:expprob}
\end{equation}
\end{figure*}

\renewcommand{\algorithmicrequire}{\textbf{Initialization:}}  
\renewcommand{\algorithmicensure}{\textbf{Process:}}
\begin{algorithm}[!t]        
\caption{The CGDT algorithm}             
\label{alg:pseu_cdt}                  
\begin{algorithmic}[1]                
\REQUIRE ~~\\                          
    $range(1) =u^{-1}(1); \ range(2)=l^{-1}(1); $
    
    $idx(1)=1; \ k=1;  $

\ENSURE ~~\\                           
    	\STATE For \ $z=2 \ \ \text{to} \ \ n$
    	\STATE \quad \quad $s=\frac{(g(z)+h^2(z))-(g(idx(k))+h^2(idx(k)))}{2h(z)-2h(idx(k))}$;
        \STATE \quad \quad Project $s$ onto interval $[u^{-1}(z),l^{-1}(z)]$;
    	\STATE \quad \quad While $s \leq range(k)$
	\STATE \quad \quad \quad \quad k = k-1;
    	\STATE \quad \quad \quad \quad $s=\frac{(g(z)+h^2(z))-(g(idx(k))+h^2(idx(k)))}{2h(z)-2h(idx(k))}$;
        \STATE \quad \quad \quad \quad Project $s$ onto interval $[u^{-1}(z),l^{-1}(z)]$;
    \STATE \quad \quad end   
    	\STATE \quad \quad If $s > range(k+1)$
    	\STATE \quad \quad \quad \quad   $k = k+1; \ \ idx(k)=z; $
    \STATE \quad \quad \quad \quad	$range(k+1)=l^{-1}(z)$;
    	\STATE \quad \quad Else
    	\STATE \quad \quad \quad \quad $k=k+1; \ \ idx(k)=z; $
    	\STATE \quad \quad \quad \quad $range(k)=s; \ \ range(k+1) = l^{-1}(z)$;
	\STATE \quad \quad end
	\STATE end
	\STATE Fill in the value of $\gamma(x)$ using $range(k)$ and $idx(k)$.
\end{algorithmic}
\end{algorithm}

The reason of making such approximation is that we can then solve problem (\ref{eq:newprob}) efficiently in linear time using the constrained generalized distance transform algorithm developed in Section \ref{sec:CDT}. We will validate this approximation by experiment in Section \ref{sec:infer}.

\subsection{Constrained Generalized Distance Transform (CGDT) Algorithm} \label{sec:CDT}
First note that since the variables $S_1 = (x_{1}, y_{1})$ are separable, we can translate the 2-dimensional problem (\ref{eq:newprob}) into two 1-dimensional problems by first minimizing one variable ($x_1$) and then minimizing the other one ($y_1$), as shown in Equation (\ref{eq:expprob}). Next we show how to efficiently solve these two similar 1-dimensional subproblems. 
To this end, we consider a slightly more general problem of the form
\begin{equation}
\gamma(x) = \min_{l(x) \leq z \leq u(x)}  {(x-h(z))^2 + g(z)}, \label{eq:coreprob}
\end{equation}
where $h(z)$, $u(x)$ and $l(x)$ are all non-decreasing. In Equation (\ref{eq:expprob}), for the inner minimization, we set $h(z) = z + \frac{\Delta x}{2}$ and $l(x)=2x-W,u(x)=2x-1$; and for the outer minimization, we set $h(z) = z + \frac{\Delta y}{2}$ and $l(y)=2y-H,u(y)=2y-1$. Note that problem (\ref{eq:coreprob}) becomes the ordinary generalized distance transform \cite{felzenszwalb2004distance} if we ignore the constraint $l(x) \leq z \leq u(x)$. Inspired by \cite{felzenszwalb2004distance}, $\gamma(x)$ can be viewed as the lower envelope of a set of truncated parabolas $(x-h(z))^2+g(z)$ with the truncation being $u^{-1}(z) \leq x \leq l^{-1}(z)$. The algorithm performs in two steps. In the first step we obtain the lower envelope of all the truncated parabolas by computing the boundary points between adjacent selected parabolas while keeping the truncation constraint being satisfied. In the second step we fill in the value $\gamma(x)$ using the obtained lower envelope from step one. Algorithm pseudocode is provided in Algorithm \ref{alg:pseu_cdt}, where we use $range(k)$ and $range(k+1)$ to indicate the range of $k$-th parabola in the lower envelope, and $idx(k)$ to indicate the grid location $z$ of the $k$-th parabola in the lower envelope.

\section{Learning for Compositional Model}
\subsection{Structure Learning} \label{sec:structlearn}
Structure learning refers to learning the hierarchical graph $\mathcal{G} = (\mathcal{V},\mathcal{E})$ to represent the animal and part shapes under various poses and viewpoints. Specifically, for each non-leaf node $v$, we need to learn the part-subpart relation $ch(v)$ and $\Delta S_v$; and for each leaf nodes $v \in \mathcal{V}_1$, we need to learn the orientation $\theta_v$ and polarity $a_v$. We consider eight orientations which are equally distributed from 0 to $\pi$, and three polarity values for each orientation which represent object region on one side, object region on the other side, and object region on both sides respectively, as shown in Figure \ref{fig:modelfigure} (b). Thus there are in total 24 types of leaf nodes at level one. Note that leaf nodes are shared across different mixtures.

We use compositional models to represent big semantic parts such as head, neck and torso, and we discuss segmenting legs in Section \ref{sec:legs}. The structure learning algorithm proceeds in the following four steps. The visualization figures are provided in the supplementary material.

1. Clustering: Given part-level annotations, we extract the masks for head, neck and torso and assign them different values (1 for head, 2 for neck, and 3 for torso). Then we resize each example by the maximal side length. We apply the K-medoids clustering algorithm to find K representative shapes from the training data. And we will build K compositional mixtures based on the K representative shapes.   

2. Sampling: We evenly sample fixed number of landmarks along the boundary of each semantic part.

3. Matching: We match each landmark to one of the 24 leaf nodes.

4. Composing: Starting from the landmarks (leaf nodes), we compose each two adjacent nodes (children) into a higher-level node (parent) and record the spatial relation between the two children nodes. The parent location is the average of two children locations. We run this procedure level-by-level up to the top level.


\subsection{Parameter Learning}\label{sec:paralearning}
The parameters of the compositional model are $w_v$ and $w_v^{\text{part}}$ for non-leaf nodes, and $w_v^{\text{edge}}$ and $w_v^{\text{app}}$ for leaf nodes. To reduce the model complexity, we assume that parameters are shared by parts. So the parameter vector becomes $\mathbf{w} = (w,w^{\text{part}},w^{\text{edge}},w^{\text{app}})$. These parameters strike a balance between the prior shape ($w$), appearance cues ($w^{\text{app}}$), orientation confidence ($w^{\text{edge}}$) and part confidence ($w^{\text{part}}$). The sharing allows us to learn the model parameters using a small number of training data. Note that the energy function $E({S_{\mathcal{V}}},\mathbf{I};\mathbf{w})$ is of the form
\begin{equation}
E({S_{\mathcal{V}}},\mathbf{I};\mathbf{w}) = \mathbf{w} \cdot \phi({S_{\mathcal{V}}},\mathbf{I}).
\end{equation}
The training dataset is denoted by $\{ (\mathbf{I}_i, y_i) \}_{i=1}^n$, where $y_i \in \{+1,-1\}$. The positive examples refer to object bounding box images and negative examples refer to bounding box images of other objects. Since we do not have the location information for all parts/subparts $S_{\mathcal{V}}$, we adopt latent SVM for learning parameters $\mathbf{w}$. 
\begin{equation}
\min_{\mathbf{w}} \quad \frac{1}{2} ||\mathbf{w}||^2+C\sum_{i=1}^n{\max(0,1-y_i F(\mathbf{I}_i;\mathbf{w}))},
\end{equation}
where the score function is defined as $F(\mathbf{I}_i;\mathbf{w}) = - \min_{S_{\mathcal{V}}}    E({S_{\mathcal{V}}},\mathbf{I}_i;\mathbf{w})$.

\section{Segmenting Legs} \label{sec:legs}
Considering the extremely high variability of animal legs, we take a coarse-to-fine approach to segment legs. Specifically, after segmenting the animal body (head, neck, torso), we can narrow down the search region for legs since we know that most of the time the legs appear underneath the torso. Then in the refined search region, we detect symmetric regions using algorithm in \cite{lee2013detecting}  since animal legs often have roughly symmetric structure. Next we compute a confidence score for each detected symmetric region $R$, and make prediction by thresholding this score. 
\begin{equation}
\text{score}(R) = w^{\text{obj}} \cdot {fea}(R).
\end{equation}
Here $w^{\text{obj}}$ is the parameters corresponding to the features extracted within object region, i.e., the first half of $w^{\text{app}}$. And $fea(R)$ is the average 34-dimensional feature vector within region $R$.

\section{Experiments} \label{sec:exp}
In this section, we will report part segmentation results for horse and cow. We also conduct some diagnostic experiments for our model. In addition, we validate by experiment that our approximate inference is much faster than exact inference while losing little accuracy.

{\bf Dataset:} We use a newly annotated dataset on Pascal VOC 2010 \cite{chendetect} to evaluate our part segmentation algorithm. It provides pixelwise semantic part annotations for each object instance. Since we focus on non-occlusion and non-truncation case, for each animal class we manually select the fully observable animal instances in both trainval and test set.  We use this refined dataset for training and testing, and we will release it. For horse and cow, there are roughly 150 fully observable bounding box images in trainval and test respectively. Considering the various poses and viewpoints of animals and cluttered background in Pascal VOC images, we believe the fully observable animal bounding box images are a suitable testbed for our algorithm.

We use the bounding box images with part annotations in the Pascal trainval set for structure learning. As for parameter learning, we use the bounding box images from the Pascal VOC trainval set as positive examples and randomly select a subset of bounding box images of other object classes from the Pascal VOC trainval set as negative examples. We use the bounding box images from the Pascal VOC test set for testing. 

{\bf Setup:} We consider head, neck, torso and leg as semantic parts. In the structure learning, we set the number of boundary landmarks to be 8 for head, 8 for neck and 16 for torso. Thus each compositional tree has 6 levels and 32 leaf nodes. The head node and neck node are at the 4-th level and torso node is at the 5-th level. We only consider the head part score (i.e., $w_v^{\text{part}}$ is non-zero only if $v$ refers to the head part) since head is the most discriminative part for animals. Our algorithm outputs the best mixture and locations of all parts/subparts. We only use the locations of all leaf nodes as boundary landmarks. We connect the adjacent leaf nodes of each semantic part to make a closed contour as part segmentation result. We use intersection-over-union (IOU) as the performance measure.

\begin{figure}[!t]
\centering	
    		\includegraphics[width=260pt]{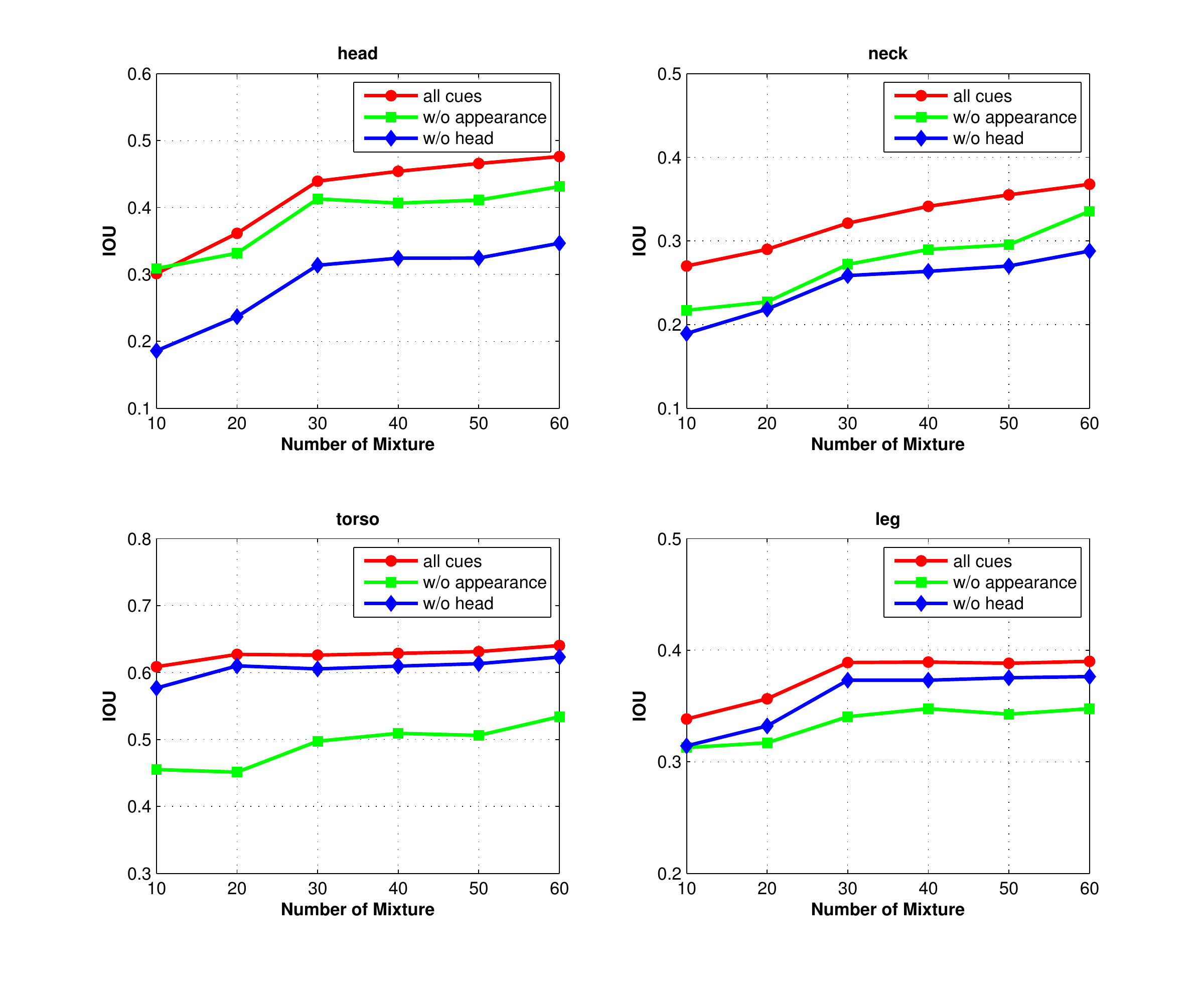}
    		\caption{The performance variation with the number of mixtures for four semantic parts. The effect of having appearance cues and head cues are also shown in this figure.} \label{fig:mixtureNum}

\end{figure}

\subsection{Efficient Inference} \label{sec:infer} Recall that we make approximations (\ref{eq:newprob}) in order to allow efficient linear-time inference. We now provide results demonstrating that we lose little in accuracy and gain much in speed by approximation. Let $E(\mathbf{I})$ denote the exact minimal energy (quadratic complexity) and $\tilde{E}(\mathbf{I})$ denote the minimal energy by our efficient approximate algorithm (linear complexity). We measure the error by $\frac{(E(\mathbf{I}) - \tilde{E}(\mathbf{I}))}{E(\mathbf{I})}$. We compute this error on all test images of horses and get 0.53\% average error. Furthermore, we compute the average location difference of all leaf nodes between exact inference and our approximate inference algorithm. We get on average 1.78 pixels error and 1.11\% if normalized by maximal side length. The above results show that our approximation is extremely accurate. As for the speed, the average parsing time per image (resize maximal side length to 160) is about 10 seconds using our inference algorithm while the average parsing time for exact inference is about 10 minutes per image. This demonstrates the significant speedup by our fast approximate inference algorithm.

\subsection{Model Diagnostics}
Our diagnostic experiment is based on horse images.

{\bf Number of Mixtures:} The structure learning algorithm uses $K$-medoids clustering to find $K$ representative shapes. Figure \ref{fig:mixtureNum} shows how the segmentation performance varies with respect to the parameter $K$ for each semantic part. Intuitively, as the number of mixture increases, our mixtures of compositional models are able to capture more deformations and variations of animal and part boundaries. Therefore, the segmentation performance improves with the number of mixtures. Particularly, small parts (head and neck) benefit significantly from increasing mixture number. 

{\bf Appearance and Head cues:} We can also see from Figure \ref{fig:mixtureNum} that the performance drops if we do not use appearance cues from semantic labeling or head cues from animal part detection. This result indicates that combining appearance and part information are necessary for localizing boundaries although they are not always correct.  

\begin{figure*}[t]
\centering
\hspace{0.4cm}
\begin{subfigure}[b]{100pt}
\centering
\includegraphics[height=100pt]{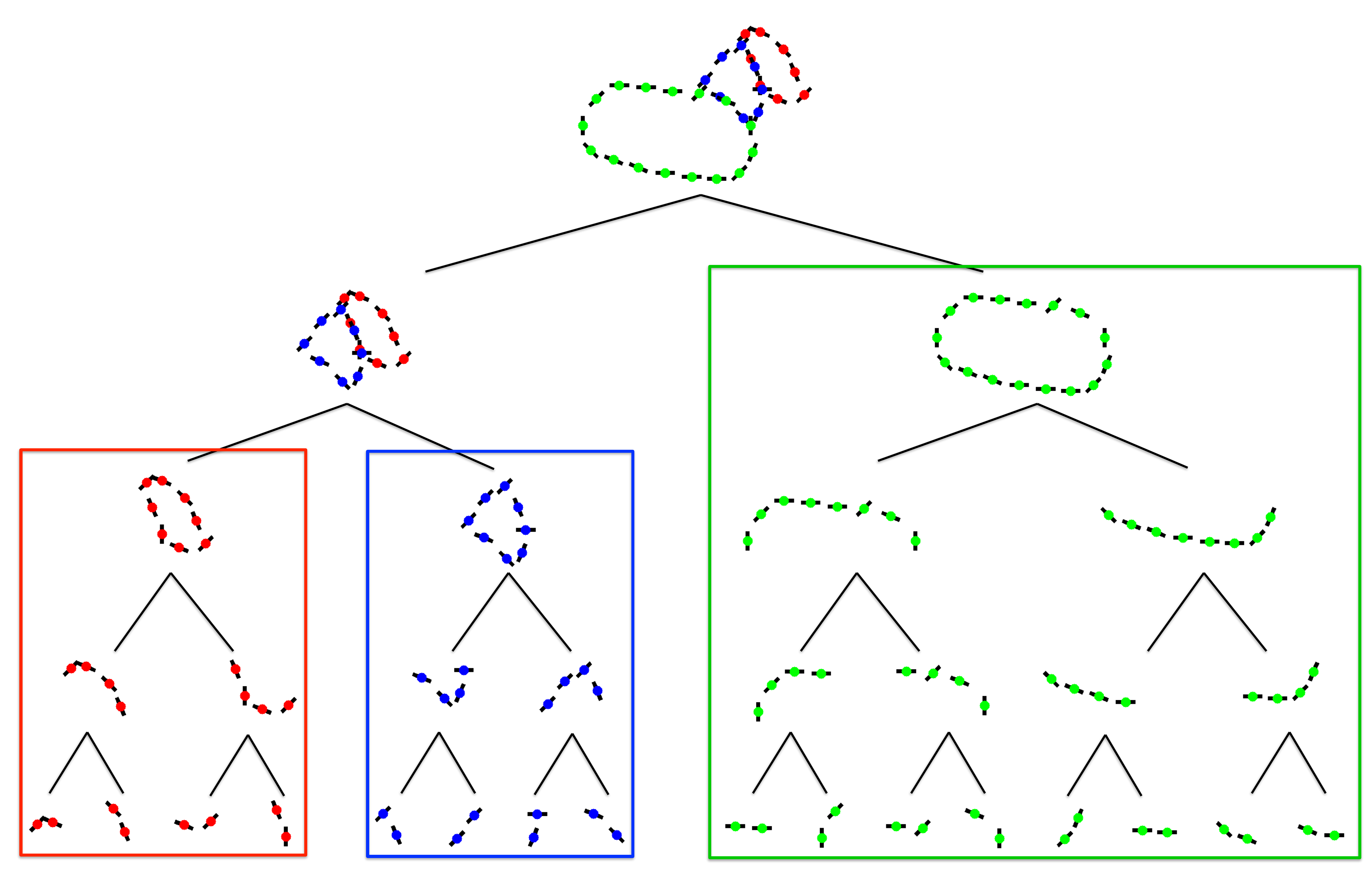}  
\end{subfigure}
\hspace{1.1cm}
\begin{subfigure}[b]{320pt}
\centering
\includegraphics[height=50pt]{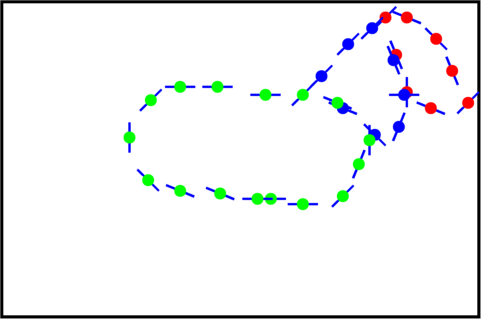}
\includegraphics[height=50pt]{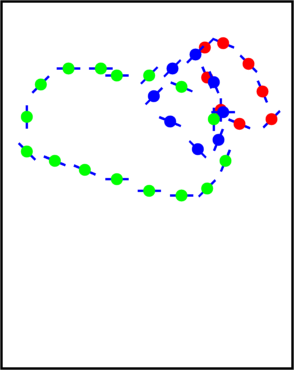}
\includegraphics[height=50pt]{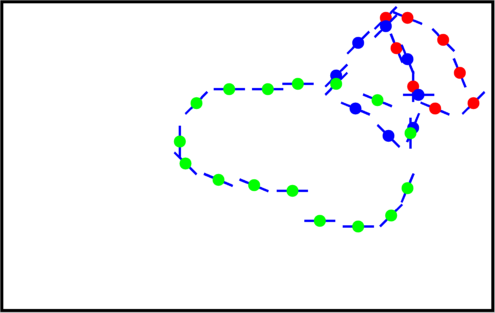}
\includegraphics[height=50pt]{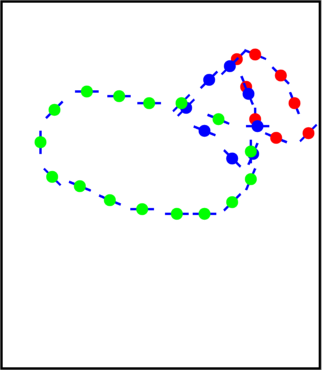}\\
\includegraphics[height=50pt]{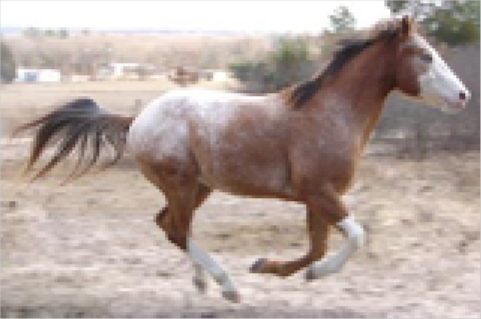}
\includegraphics[height=50pt]{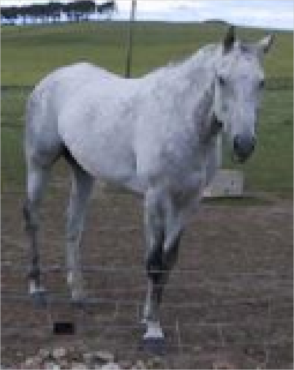}
\includegraphics[height=50pt]{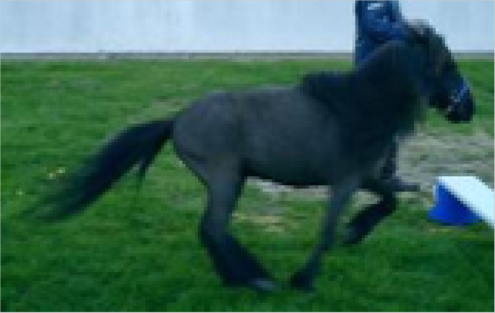}
\includegraphics[height=50pt]{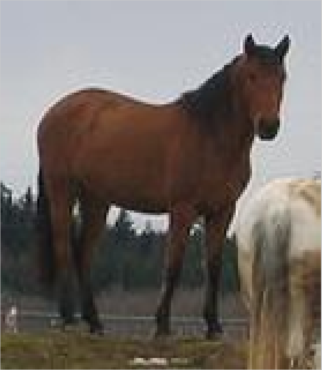}
\end{subfigure} 

\vspace{0.2cm}
\hspace{0.5cm}
\begin{subfigure}[b]{100pt}
\centering
\includegraphics[height=100pt]{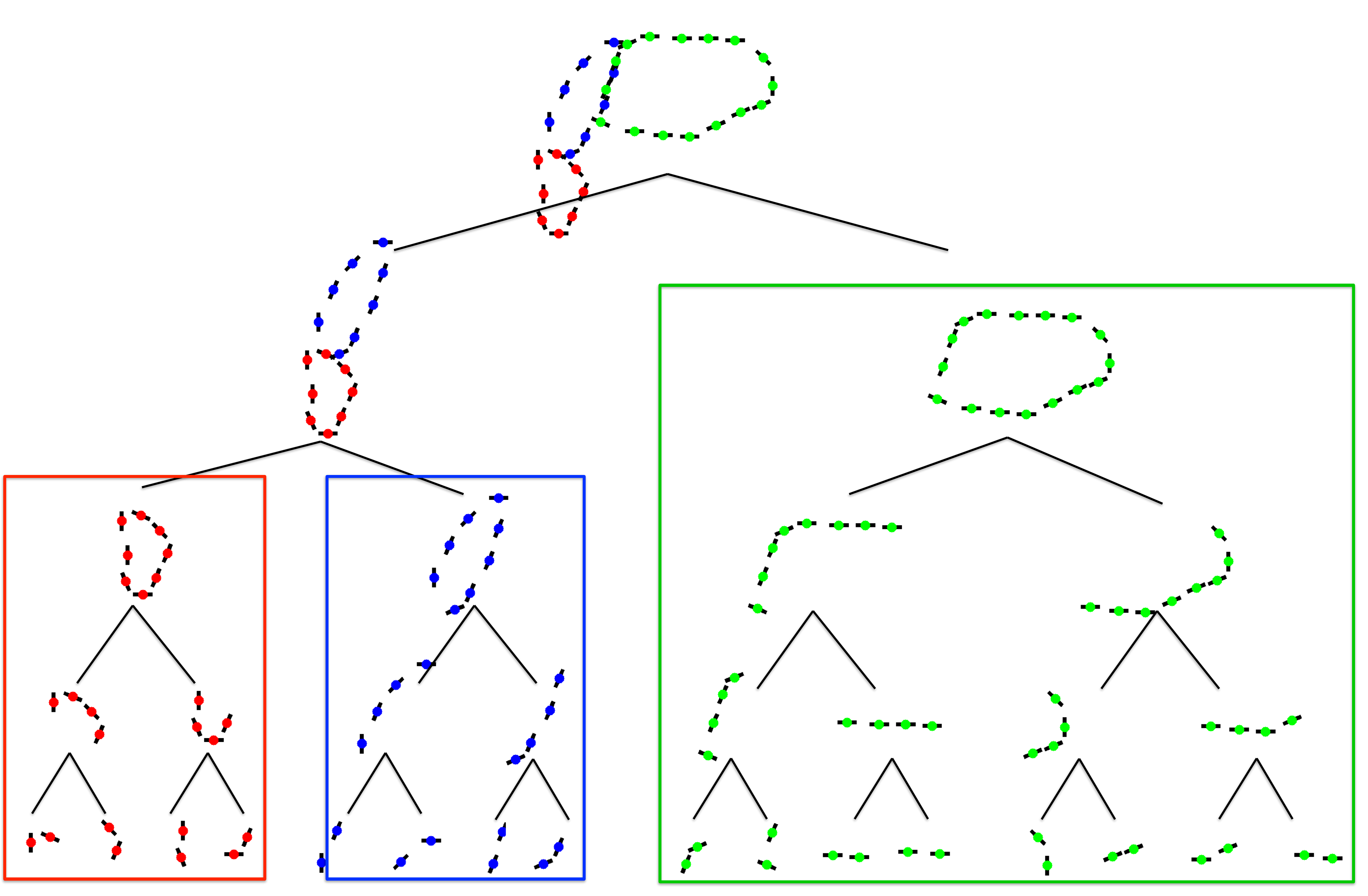}
\end{subfigure}
\hspace{1.0cm}
\begin{subfigure}[b]{320pt}
\centering
\includegraphics[height=50pt]{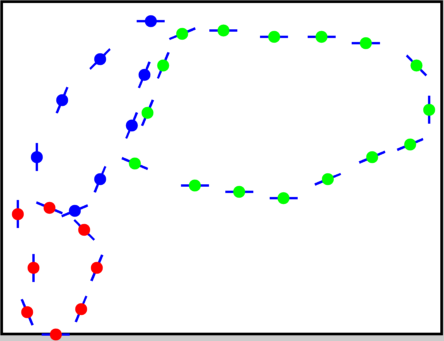}
\includegraphics[height=50pt]{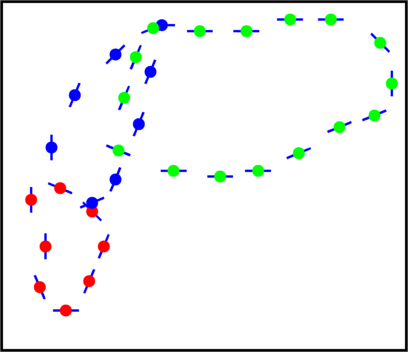}
\includegraphics[height=50pt]{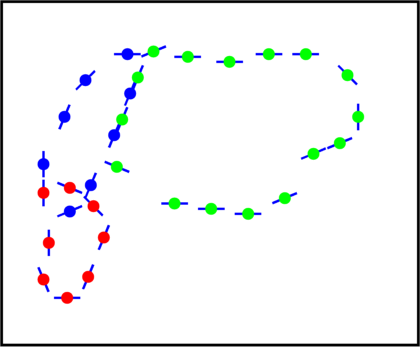}
\includegraphics[height=50pt]{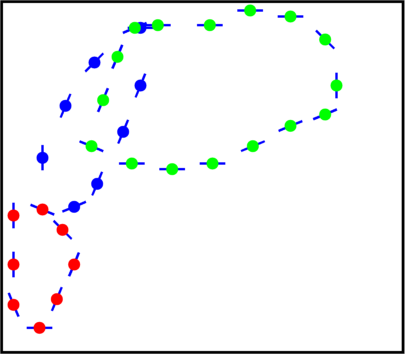} \\ 
\includegraphics[height=50pt]{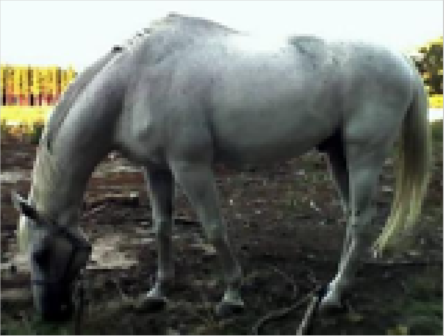}
\includegraphics[height=50pt]{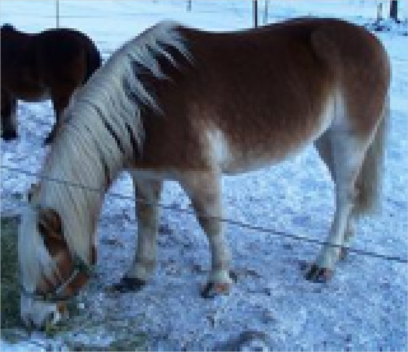}
\includegraphics[height=50pt]{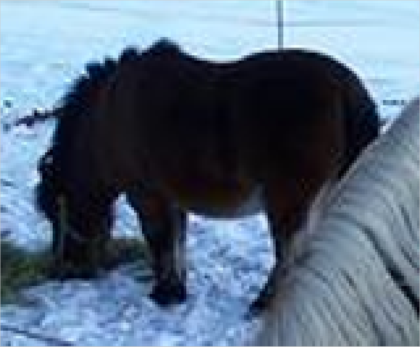}
\includegraphics[height=50pt]{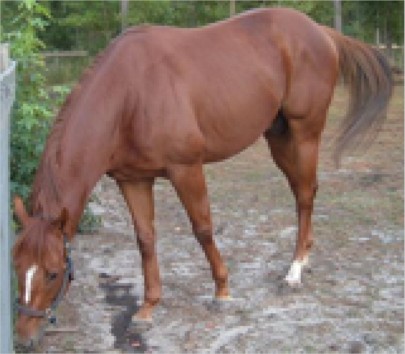} 
\end{subfigure} 
\caption{Two mixtures and corresponding landmark localization results. For each mixture, the left figure is the compositional model, the top row on the right is the landmark localization results, and the bottom row on the right is the input images. We can see that each mixture deals with local deformation, and different mixtures handle large variation due to poses and viewpoints.}
\label{fig:deform}
\end{figure*}

{\bf Deformation Ability of Compositional Model:} Figure \ref{fig:deform} shows that each mixture deals with local deformation, and different mixtures handle large global variation due to poses and viewpoints. Thus our mixtures of compositional models are able to capture various shape variations.

{\bf Failure Cases:}
Figure \ref{fig:failure} shows three typical failure cases. The reason for (a) is because the horse is in very rare pose which cannot be captured by any mixture. The reason for (b) is because the semantic labeling result is wrong and the horse boundary is unclear due to dark lighting. Incorrect body segment often leads to wrong leg segmentation (e.g., case (a) and (b)). In (c), the legs are mistakenly segmented although the horse body segment is correct. This is because both the detected symmetric structure (red region on the image) and the semantic labeling result are not correct.

\begin{table}[t] 
\centering
\begin{tabular}{l*{6}{c}r} 
\hline
Method                &head & neck+torso & leg\\
\hline
Our model             &34.82   & 55.62 &28.56\\
PD+OS                 &26.77 & 53.79 & 11.18 \\
\hline
PD+GT				 &38.66 & 60.63 & 19.36 \\
\hline
\end{tabular} 

\vspace{0.3cm}
\begin{tabular}{l*{6}{c}r} 
\hline
Method                &head & neck & torso & neck+torso & leg\\
\hline
Our model             &47.21 & 38.01 & 61.02 & 66.74 & 38.18\\
PD+OS                 &37.32 & N/A & N/A & 60.35 & 27.47 \\
\hline
PD+GT                 &56.64 & N/A & N/A & 67.96 & 40.95 \\
\hline
\end{tabular} 
\caption {Part segmentation result for horses (bottom) and cows (top). The performance measure is IOU (\%). PD+OS refers to the method that combines part detection bounding box and object segmentation. PD+GT refers to the method that combines part detection bounding box and groundtruth segmentation.}\label{tab:horseresult}
\end{table}

\subsection{Comparison}
{\bf Baseline:} There has been lack of work on semantic part segmentation for animals. But there is part-based object detection work \cite{chendetect} that is able to output part-level bounding boxes. There is also many object segmentation works that give object-level segments. Therefore, it is straightforward to combine part detection and object segmentation to output part-level segmentation result. Take the head as an example. We treat certain part of the object segment that lies inside the head bounding box as the head segment. This method is our comparison baseline. We use the state-of-the-art object segmentation algorithm \cite{BharathECCV2014} in the experiment.

\begin{figure*}[t]
\centering
\begin{subfigure}[b]{450pt}
\includegraphics[height=50pt]{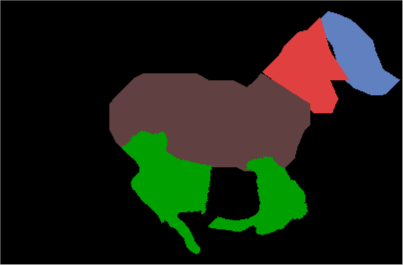}
\includegraphics[height=50pt]{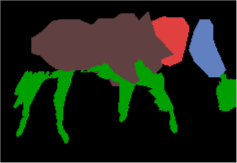}
\includegraphics[height=50pt]{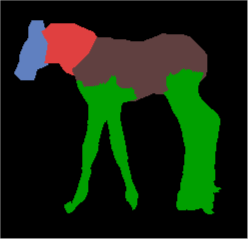}
\includegraphics[height=50pt]{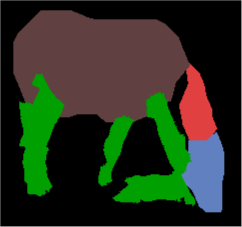}
\includegraphics[height=50pt]{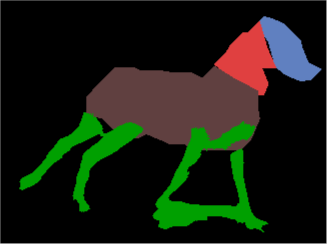}
\includegraphics[height=50pt]{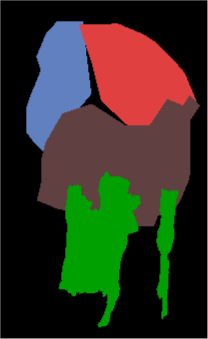} 
\includegraphics[height=50pt]{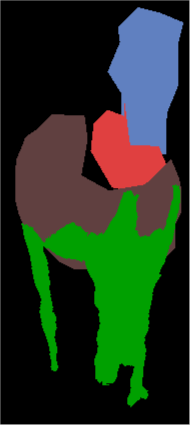}
\includegraphics[height=50pt]{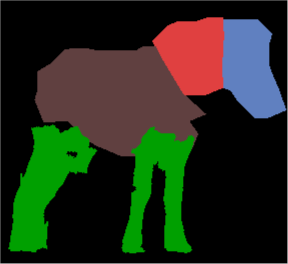}

\includegraphics[height=50pt]{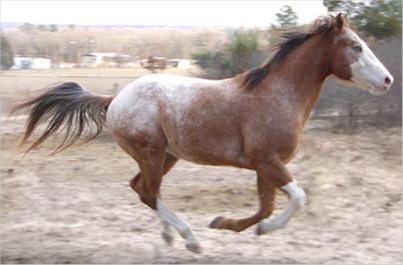}
\includegraphics[height=50pt]{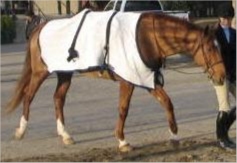}
\includegraphics[height=50pt]{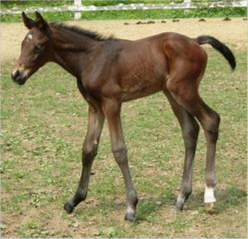}
\includegraphics[height=50pt]{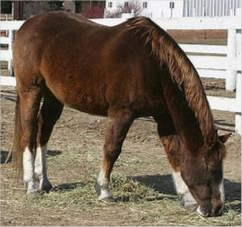}
\includegraphics[height=50pt]{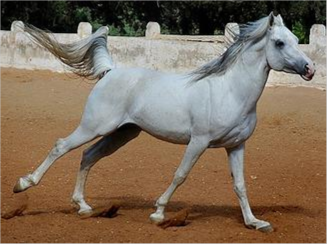}
\includegraphics[height=50pt]{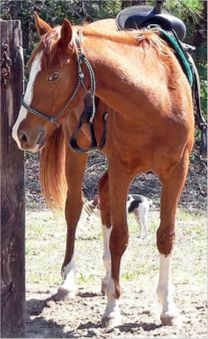} 
\includegraphics[height=50pt]{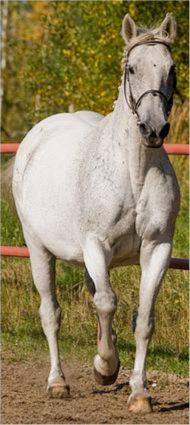}
\includegraphics[height=50pt]{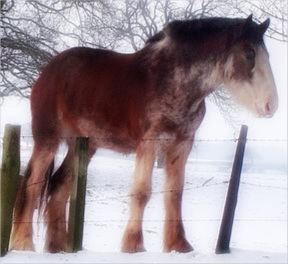}
\end{subfigure}
\begin{subfigure}[b]{10pt}
\centering
\includegraphics[height=100pt]{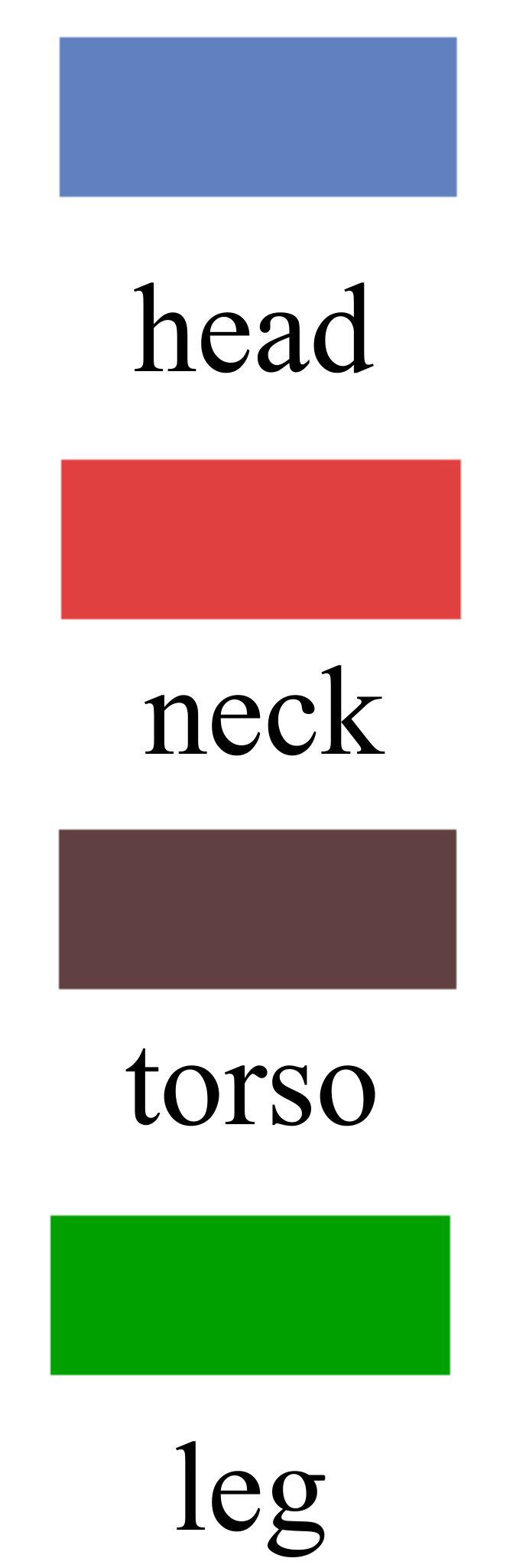}
\end{subfigure}

\vspace{0.2cm}
\begin{subfigure}[b]{450pt}
\includegraphics[height=50pt]{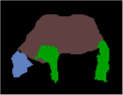}
\hspace{0.03cm}
\includegraphics[height=50pt]{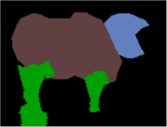}
\hspace{0.03cm}
\includegraphics[height=50pt]{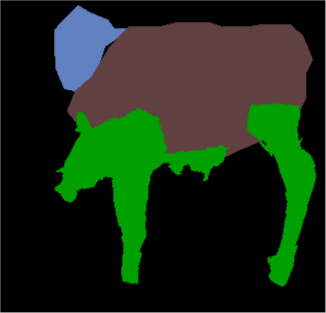}
\hspace{0.03cm}
\includegraphics[height=50pt]{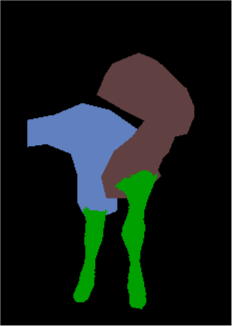}
\hspace{0.03cm}
\includegraphics[height=50pt]{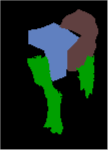}
\hspace{0.03cm}
\includegraphics[height=50pt]{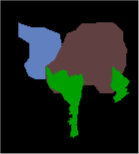}
\hspace{0.03cm}
\includegraphics[height=50pt]{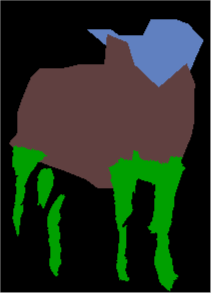}
\hspace{0.03cm}
\includegraphics[height=50pt]{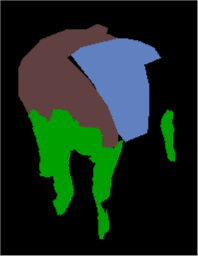}
\hspace{0.03cm}
\includegraphics[height=50pt]{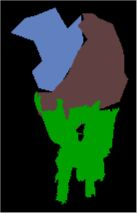}

\includegraphics[height=50pt]{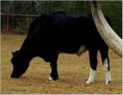}
\hspace{0.03cm}
\includegraphics[height=50pt]{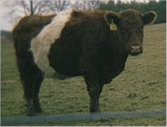}
\hspace{0.03cm}
\includegraphics[height=50pt]{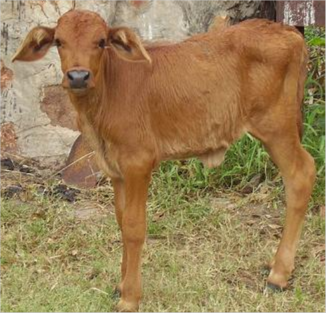}
\hspace{0.03cm}
\includegraphics[height=50pt]{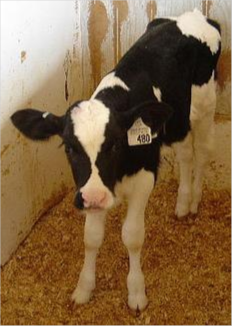}
\hspace{0.03cm}
\includegraphics[height=50pt]{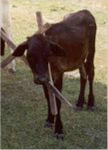}
\hspace{0.03cm}
\includegraphics[height=50pt]{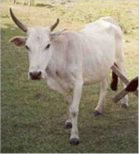}
\hspace{0.03cm} 
\includegraphics[height=50pt]{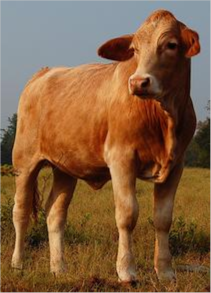}
\hspace{0.03cm}
\includegraphics[height=50pt]{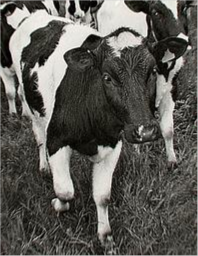}
\hspace{0.03cm}
\includegraphics[height=50pt]{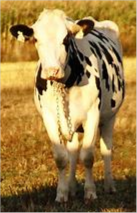}
\end{subfigure}
\begin{subfigure}[t]{10pt}
\centering
\includegraphics[height=80pt]{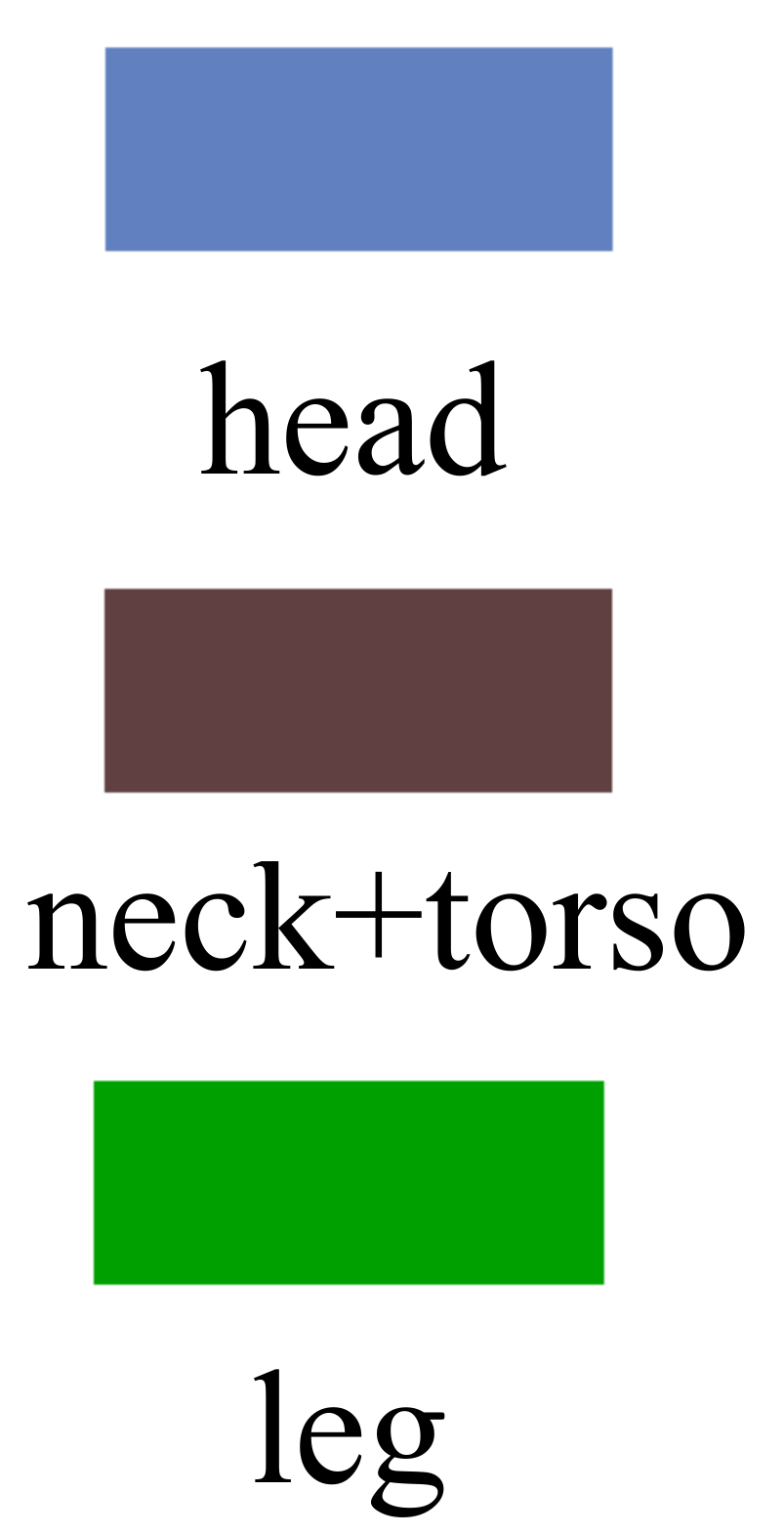}
\end{subfigure}
\caption{Typical semantic part segmentation results from various viewpoints and poses for horses (top) and cows (bottom). Best viewed in color.}
\label{fig:visual}
\end{figure*}

\begin{figure*}[!t]
\begin{subfigure}[b]{170pt}
\includegraphics[height=45pt]{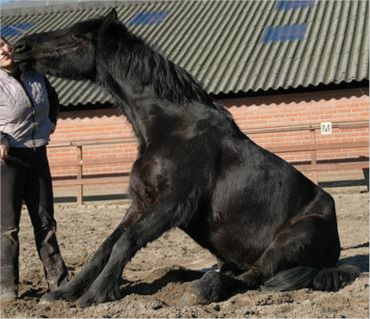}
\includegraphics[height=45pt]{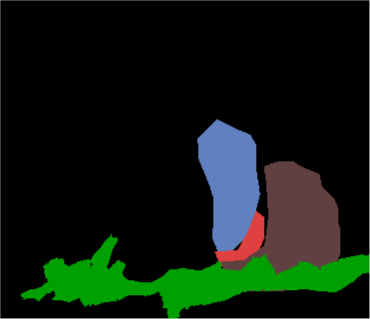}
\includegraphics[height=45pt]{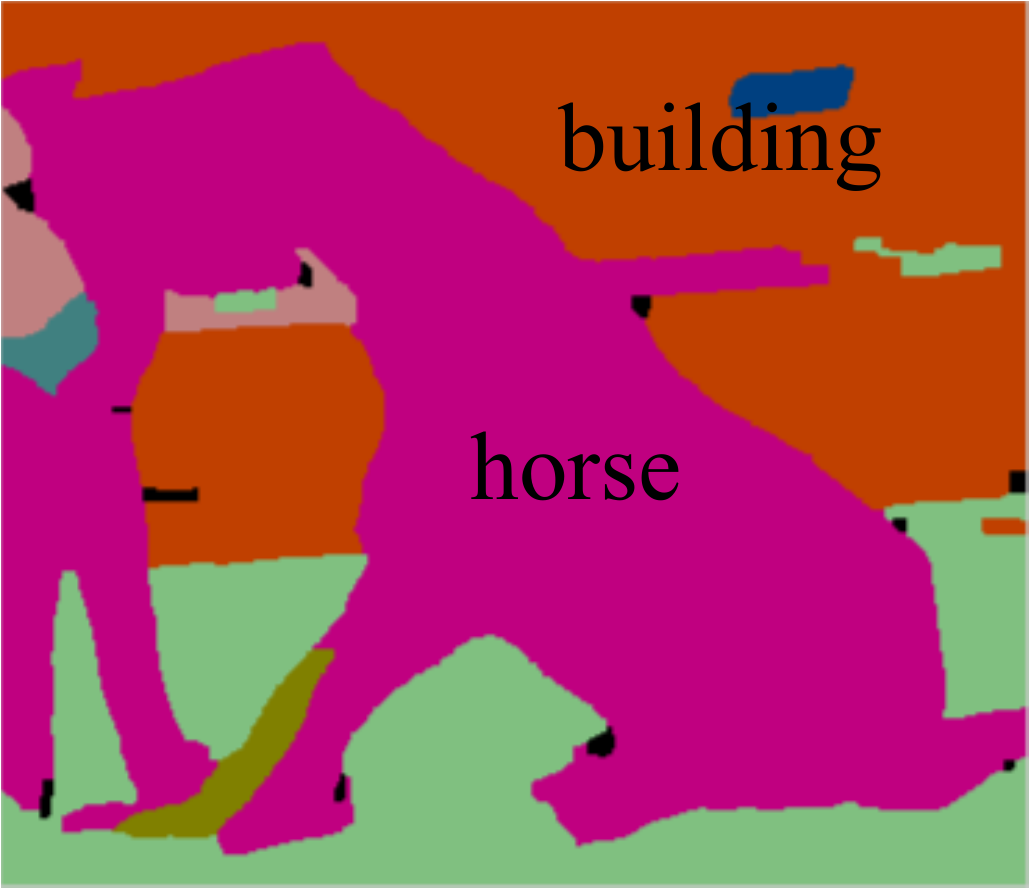}
\caption{}
\end{subfigure}
\hspace{0.01cm}
\begin{subfigure}[b]{160pt}
\includegraphics[height=45pt]{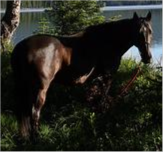}
\includegraphics[height=45pt]{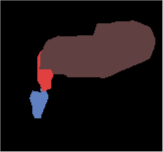}
\includegraphics[height=45pt]{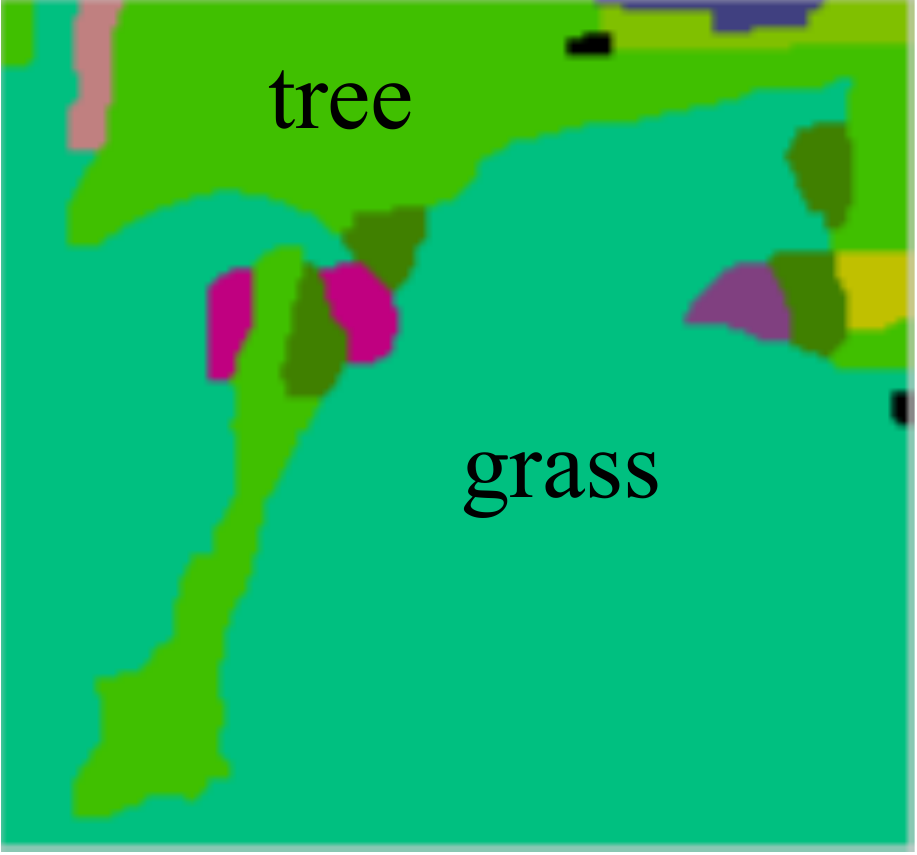}
\caption{}
\end{subfigure}
\begin{subfigure}[b]{170pt}
\includegraphics[height=45pt]{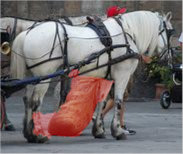}
\includegraphics[height=45pt]{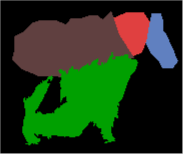}
\includegraphics[height=45pt]{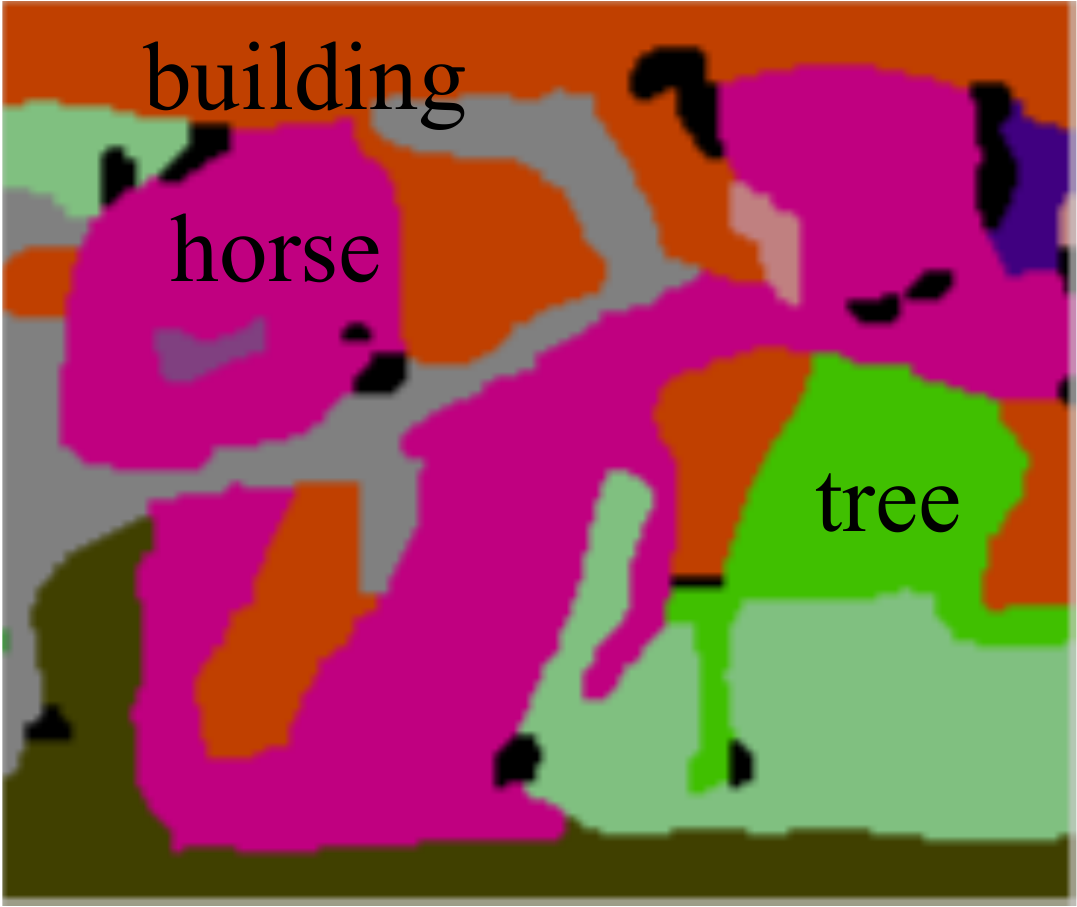}
\caption{}
\end{subfigure}
\caption{Three typical failure cases. For each case, left is image, middle is part segmentation result, and right the semantic labeling result. (a) rare pose. (b) mistaken semantic labeling and unclear boundary. (c) correct body segment but wrong leg segment due to mistaken semantic labeling and symmetric structure (red region on the image).}
\label{fig:failure}
\end{figure*}

We conduct our experiments on two animal classes: horse and cow. Table \ref{tab:horseresult} shows quantitative results and Figure \ref{fig:visual} gives some part segmentation visualizations. The horse model has 60 mixtures and cow model has 30 mixtures. Since the part detection method \cite{chendetect} treats neck+torso as one part, we do not have detection bounding box for neck and torso separately. For cows we did not split neck and torso since the cow neck is always small, which is in contrast to the long horse neck. We can see that our part segmentation results are significantly better than the baseline methods (PD+OS). We can also see that our results are only a little lower than the method (PD+GT) that combines the part detection bounding box and the groundtruth animal segmentation. Note that this is an "oracle" method since groundtruth segmentation is never available during test time. This result further validates the effectiveness of our method. 

\section{Conclusion}
In this paper, we built a mixture of compositional models combining shape and appearance for animal part segmentation task. We proposed a novel structure learning algorithm to learn the mixture of compositional trees which are able to represent animal shapes of various poses and viewpoints. We also developed a linear complexity algorithm to significantly speed up the inference of the compositional model. We tested our method for horse and cow on the Pascal VOC dataset. The experimental results showed that our method achieves much better part segmentation results than the baseline method. As for the future work, we will deal with occlusion and truncation issue, and enable part sharing when learning the compositional models.

\newpage

\section*{Supplementary Material}

\section*{A. Effect of Increasing Training Data}
\begin{figure}[!h]
\centering
\includegraphics[width = 0.45\textwidth]{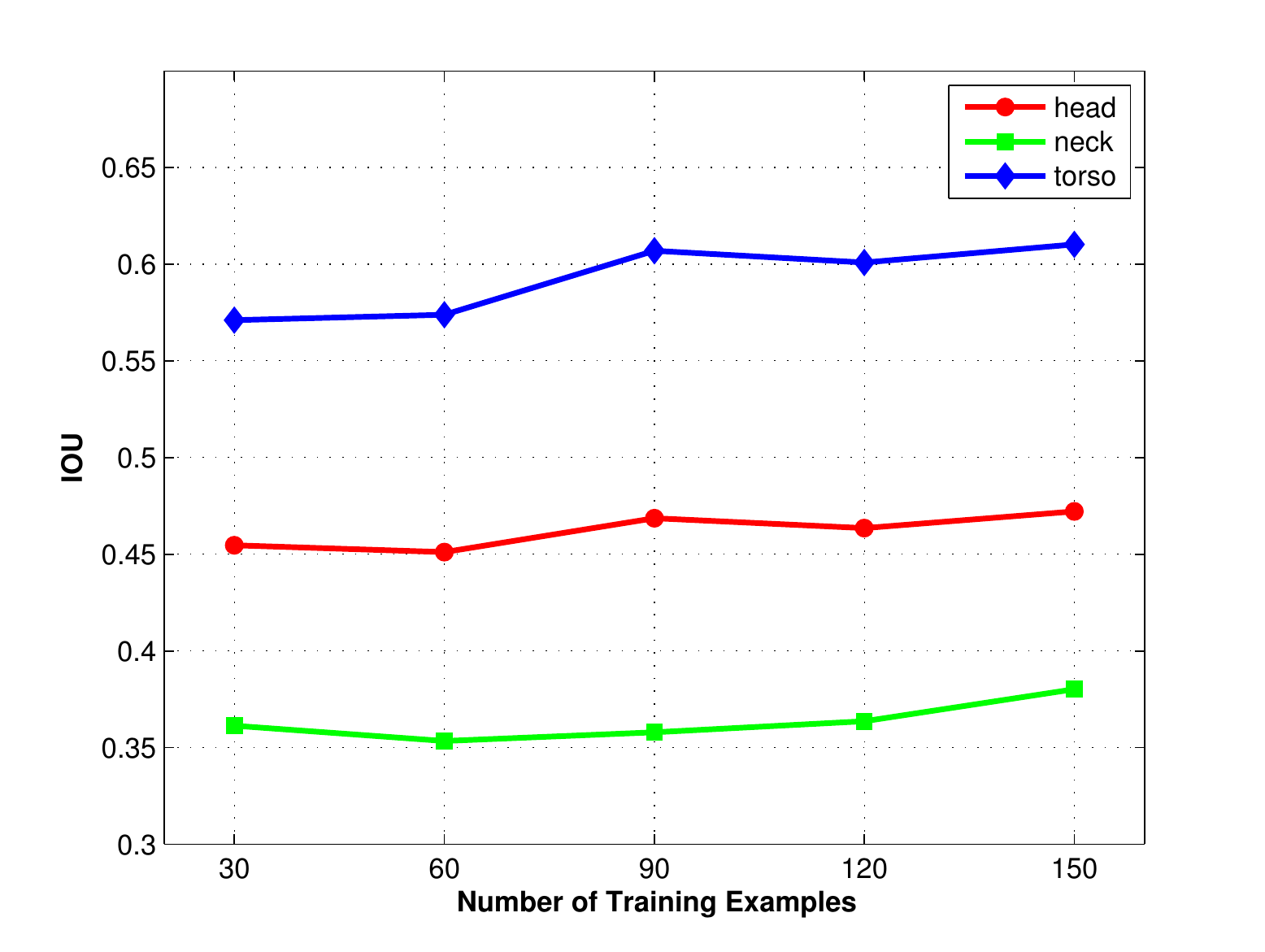}
\caption{The segmentation performance with respect to the number of training images.}
\label{fig:supple}
\end{figure}

Our model has a small number of parameters due to parameter sharing across parts, which enables to learn the model parameters using limited training data. Figure \ref{fig:supple} shows that the segmentation performance only slightly increases with respect to the number of training images. We can see that our model performs very well even using 30 training bounding box images. This indicates that our compositional model is able to learn the model parameters using very limited number of training examples, which we think is another advantage of our model.

\section*{B. Visualization of Structure Learning Algorithm}
\begin{figure}[!t]
\centering
\includegraphics[width = 0.45\textwidth]{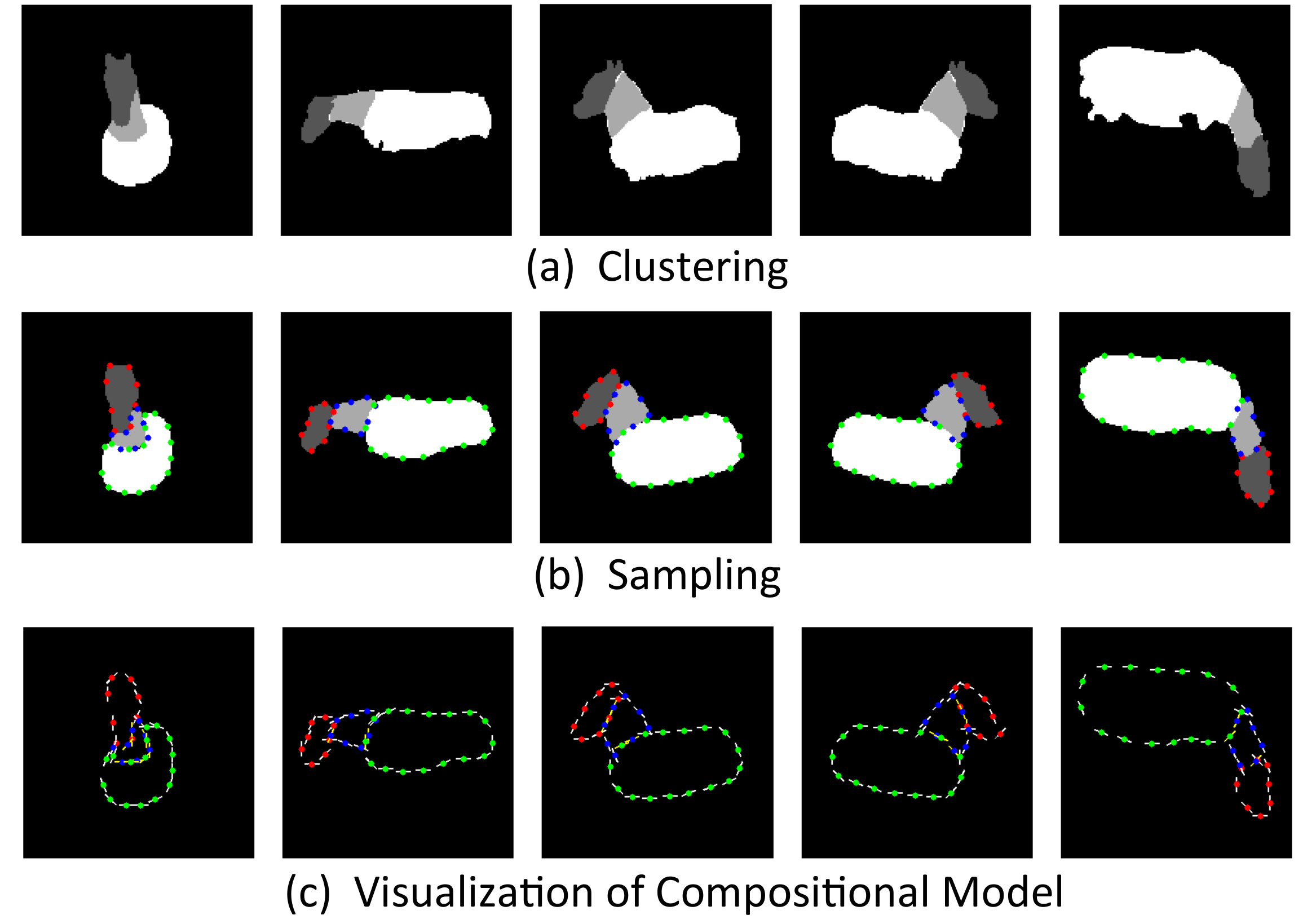}
\caption{Visualization of results from each step of the structure learning algorithm. Red for head, blue for neck, and green for torso. In (c), the line segment refers to the oriented edge. Best viewed in color.}
\label{fig:structlearn}
\end{figure}

The structure learning algorithm in Section 5.1 of the paper includes four steps: clustering, sampling, matching, and composing. Figure \ref{fig:structlearn} shows the intermediate results from clustering, sampling and composing (final compositional model visualized in a flat manner).

\begin{figure*}[!t]
\centering
\includegraphics[width = \textwidth]{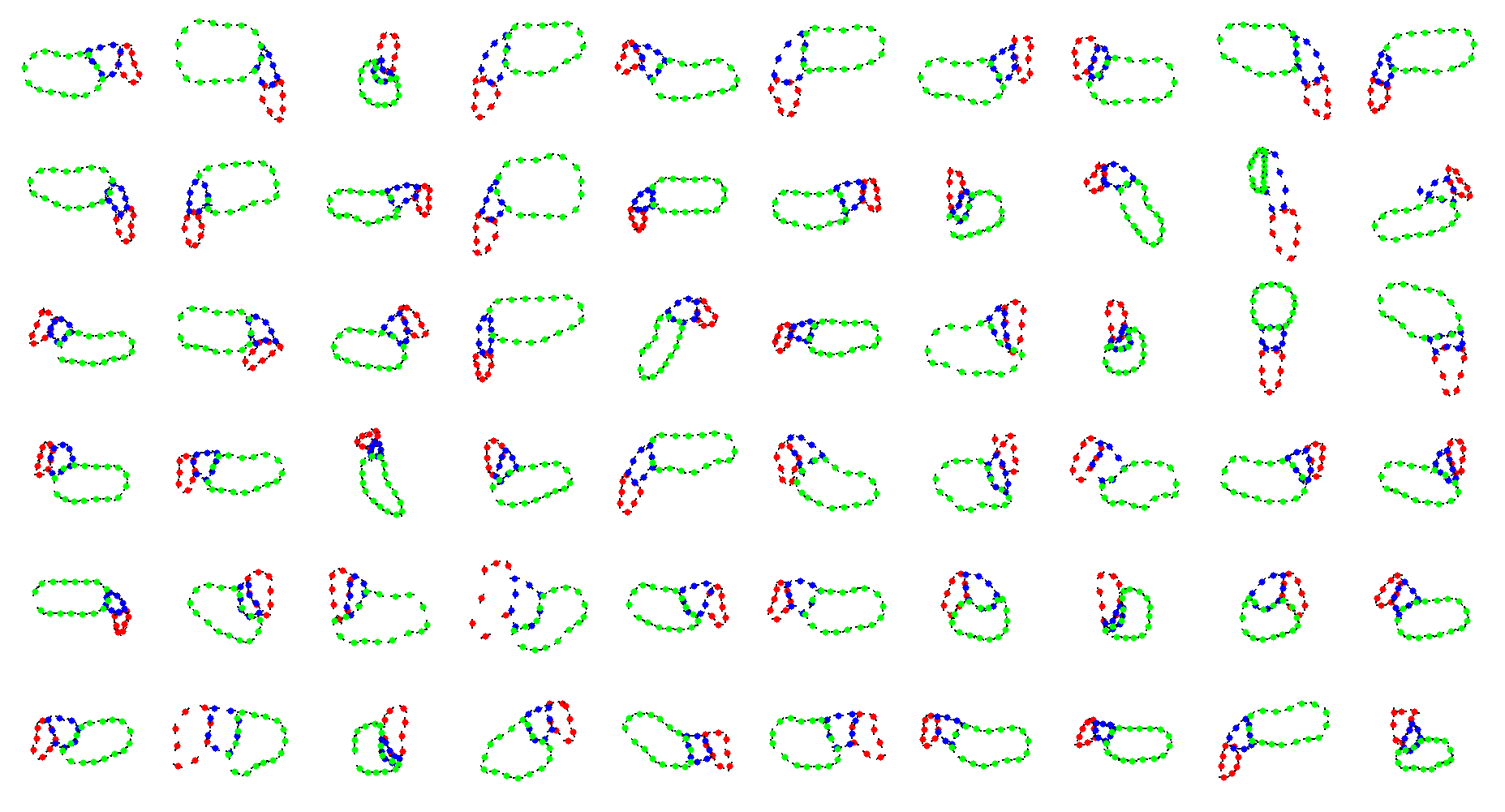}
\caption{Visualization of 60 mixtures for horses. Red for head, blue for neck, and green for torso. Best viewed in color.}
\label{fig:mixture60}
\end{figure*}
\section*{C. Visualization of Compositional Models}
Figure \ref{fig:mixture60} shows all 60 mixtures we used for horse images. Due to space limitation, the shapes are visualized in a flat manner. The hierarchical visualizations for the compositional models are shown in Figure 2 and Figure 4 of the paper.

\section*{D. Proof of Algorithm 1}

In this section, we provide a brief proof for the Algorithm 1 in the paper.
We consider the following problem 
\begin{equation*}
\gamma(x) = \min_{l(x) \leq z \leq u(x)}  {(x-h(z))^2 + g(z)},
\end{equation*}
where $h(z)$, $u(x)$ and $l(x)$ are all non-decreasing. The variables $x$ and $z$ are defined on a 1-dimensional grid $\{1,2,...,n\}$. Inspired by \cite{felzenszwalb2004distance}, $\gamma(x)$ can be viewed as the lower envelope of a set of truncated parabolas $(x-h(z))^2+g(z)$ with the truncation being $u^{-1}(z) \leq x \leq l^{-1}(z)$. The algorithm performs in two steps. The first step is that we obtain the lower envelope of all the truncated parabolas by computing the boundary points between adjacent selected parabolas while keeping the truncation constraint being satisfied. The second step is that we fill in the value $\gamma(x)$ using the obtained lower envelope from step one. In the paper, we use $range(k)$ and $range(k+1)$ to indicate the range of $k$-th parabola in the lower envelope, and $idx(k)$ to indicate the grid location $z$ of the $k$-th parabola in the lower envelope. In the proof, for notational simplicity, we use $r(k)$ to refer to $range(k)$ and $i(k)$ to refer to $idx(k)$. 
\begin{figure}[!h]
\centering
\includegraphics[width = 0.45\textwidth]{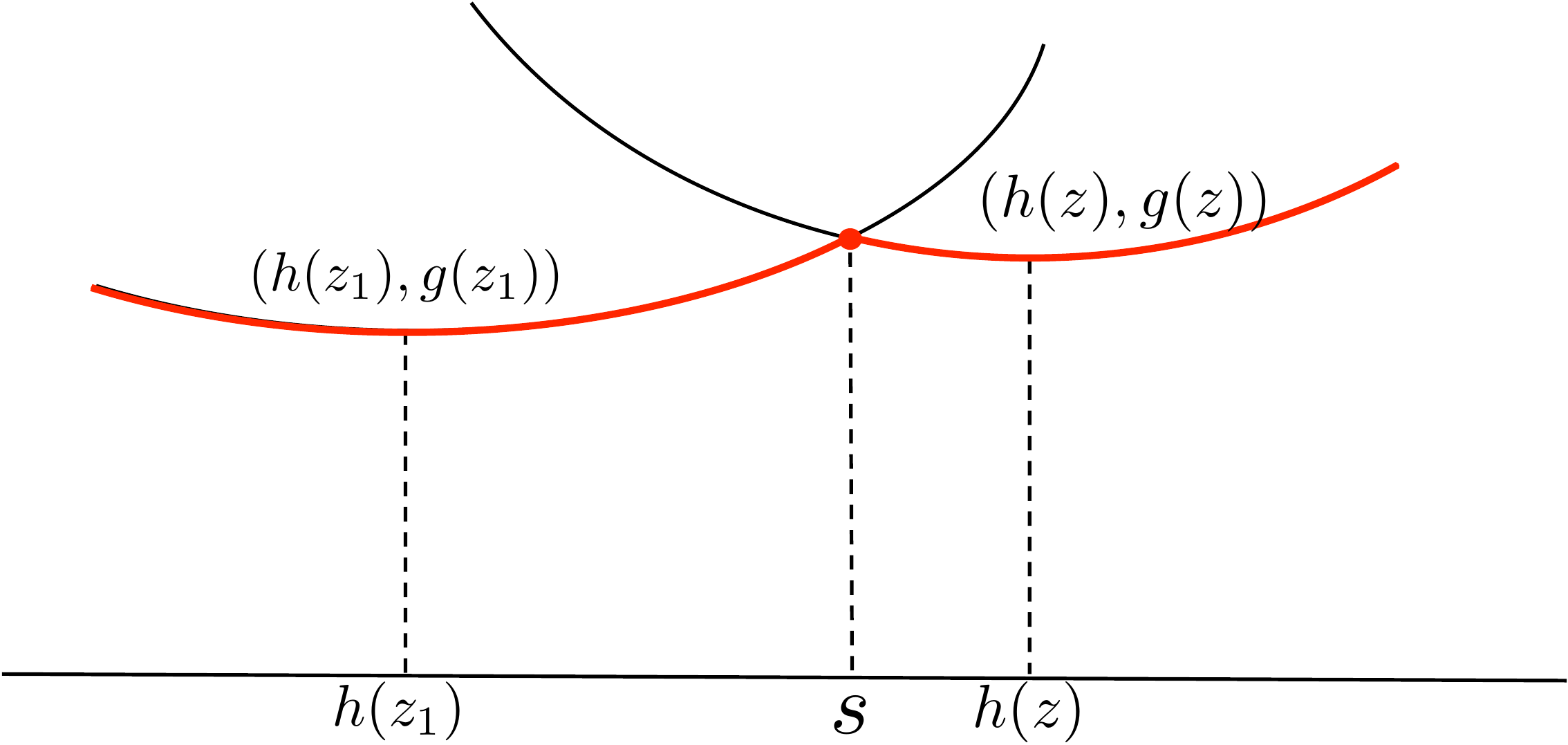}
\caption{The lower envelope computation for two parabolas.} 
\label{fig:proof_start}
\end{figure}	

As shown in Figure \ref{fig:proof_start}, the lower envelope computation for two parabolas is as follows. We first compute the intersection point 
\begin{equation}
s = \frac{(g(z)+h^2(z))-(g(z_1)+h^2(z_1))}{2h(z)-2h(z_1)}.
\end{equation}
For $x \leq s$, the lower envelope is the left parabola rooted at $(h(z_1),g(z_1))$; for $s>x$, the lower envelope is the right parabola rooted at $(h(z),g(z))$.

The algorithm performs with $z$ being from $1$ to $n$. Each time we check the parabola rooted at $(h(z),g(z))$, and update the lower envelope set accordingly. Now suppose there are already $k$ parabolas selected in the lower envelope set. For a new value $z$, we compute the lower envelope  between the parabola rooted at $(h(z),g(z))$ and the rightmost parabola in the lower envelope set rooted at $(h(i(k),g(i(k))))$. We can easily compute their intersection 
\begin{equation}
s = \frac{(g(z)+h^2(z))-(g(i(k))+h^2(i(k)))} {2h(z)-2h(i(k)) }.
\end{equation}
To satisfy the truncation constraint, we project $s$ to interval $[u^{-1}(z), \ l^{-1}(z)]$ . We consider the following three cases for computing the boundary points. We use $s^*$ to denote the projected $s$.

\begin{figure}[h]
\centering
\includegraphics[width = 0.45\textwidth]{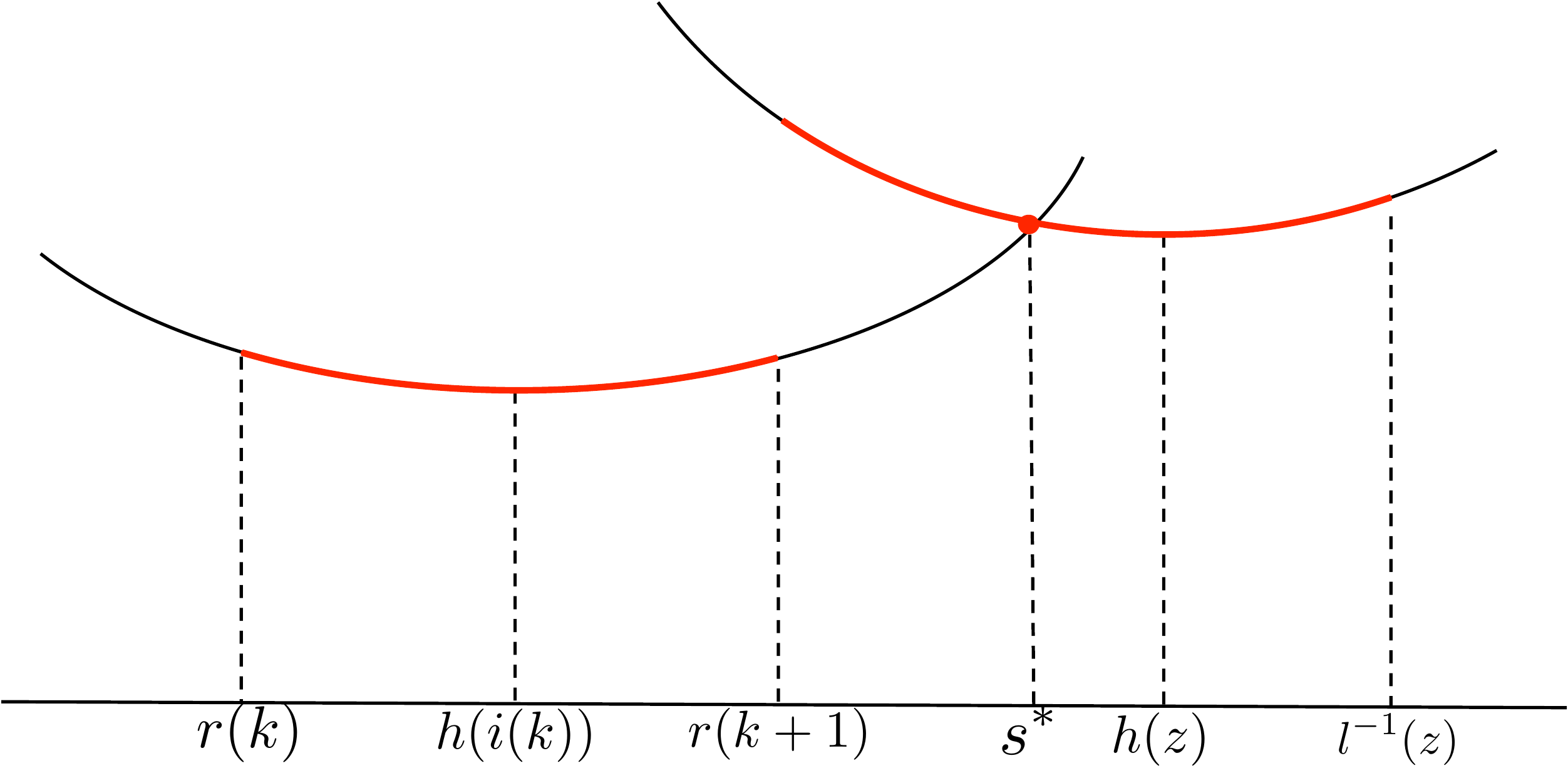}
\caption{Case 1. Best viewed in color.} 
\label{fig:proof_case1}
\end{figure}

\begin{figure}[h]
\centering
\includegraphics[width = 0.45\textwidth]{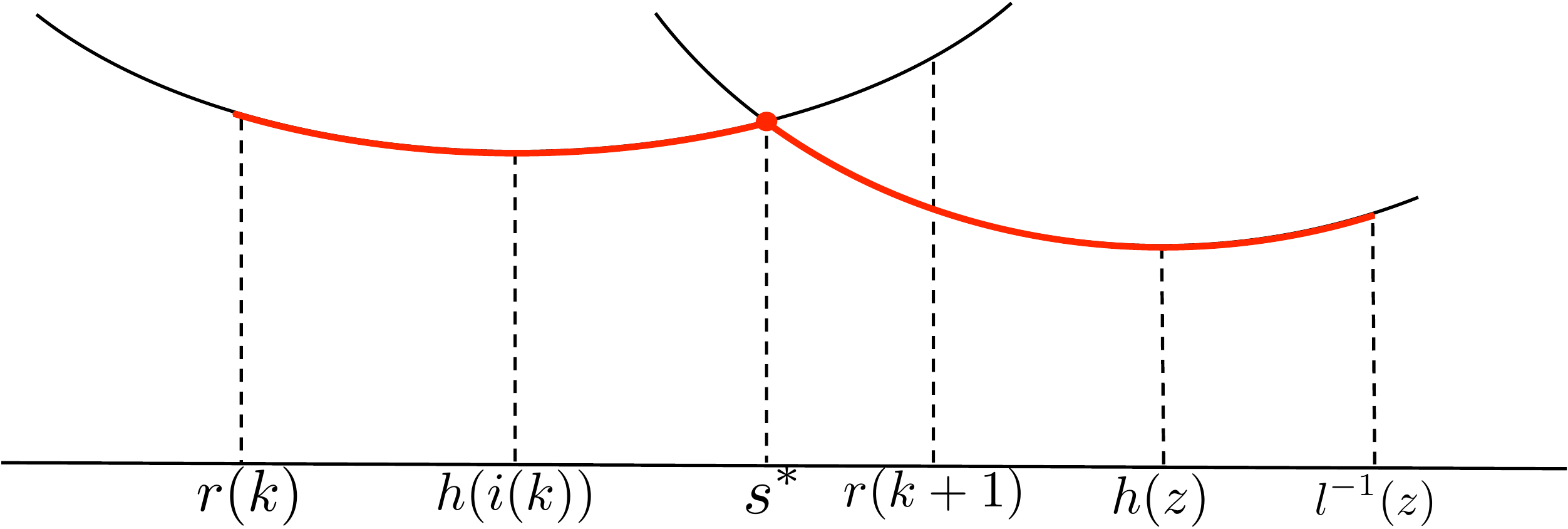}
\caption{Case 2. Best viewed in color.} 
\label{fig:proof_case2}
\end{figure}

\begin{figure}
\centering
\includegraphics[width = 0.45\textwidth]{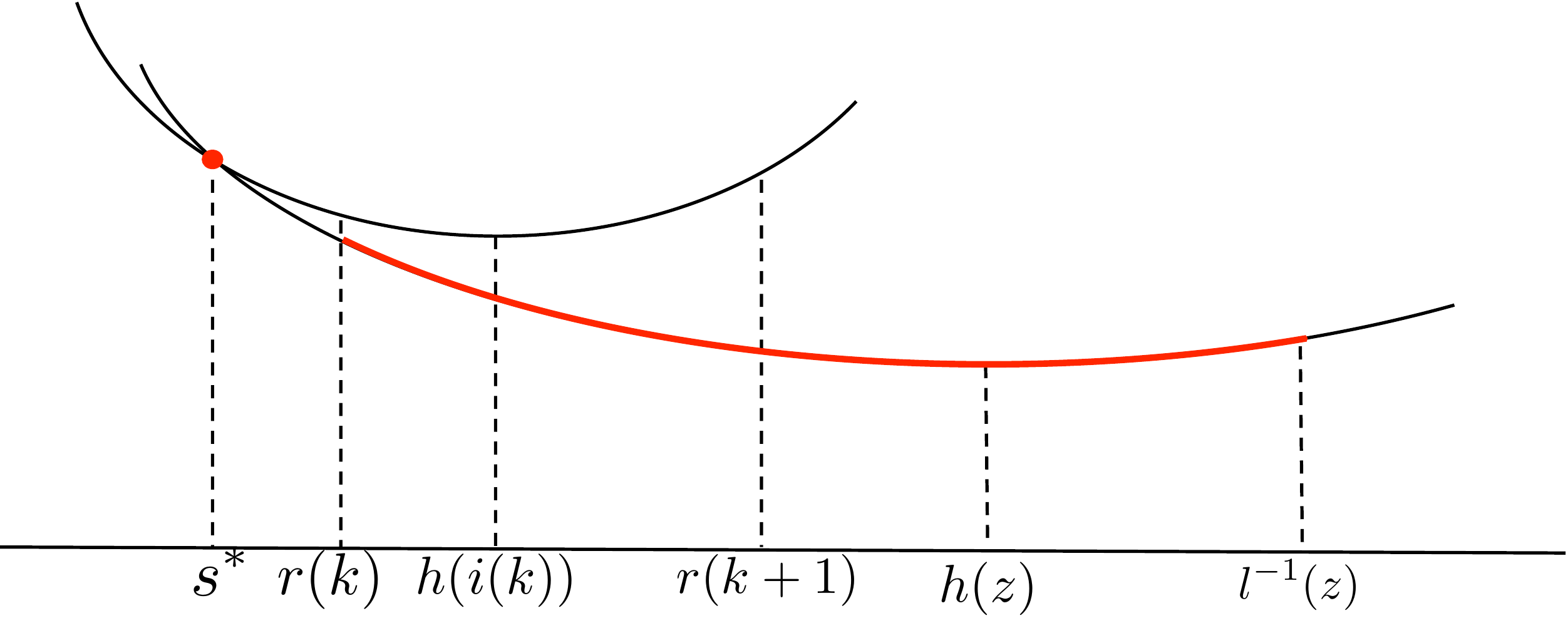}
\caption{Case 3. Best viewed in color.} 
\label{fig:proof_case3}
\end{figure}

Figure \ref{fig:proof_case1}  shows the first case where $s^* > r(k+1)$. Note that the Algorithm 1 in the paper automatically implies that
\begin{equation}
r(k+1) = l^{-1}(i(k)).
\end{equation}
In this case, we add the current parabola induced by $h(z)$ to the lower envelope set. Its range is $[r(k+1),l^{-1}(z)]$. And we do not need to consider the other parabolas in the envelope set.

Figure \ref{fig:proof_case2} shows the second case where $s^* \in [r(k),r(k+1)]$. In this case, we add the current parabola induced by $h(z)$ to the lower envelope set. And its range is $[s,l^{-1}(z)]$. As with the first case, we do not need to consider the other parabolas in the envelope set either.

Figure \ref{fig:proof_case3} shows the third case where $s^* < r(k)$. In this case, we can only guarantee that for the parabola induced by $h(z)$, the range $[r(k),l^{-1}(z)]$ is definitely in the lower envelop set. This means that we remove the $k$-th parabola (induced by h(i(k))) in the lower envelope. But for the $x<r(k)$, we have to compare the parabola induced by $(h(z),g(z))$ with other parabolas in the lower envelope set by iteratively decreasing $k$. And for each new $k$, we repeat the same operations discussed above.

\begin{figure}[h]
\centering
    		\includegraphics[width= 0.45\textwidth]{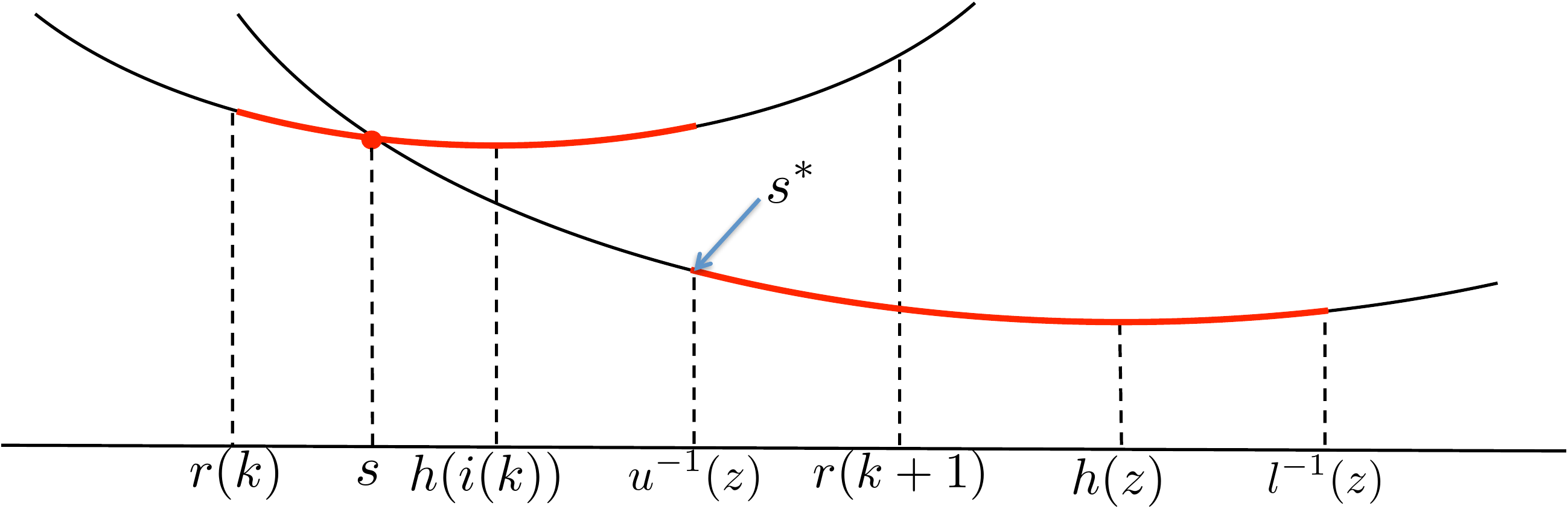}
    		\caption{ Discontinuous boundary. Best viewed in color.} \label{fig:proof_case4}

\end{figure}

Note that each parabola is at most removed once from the lower envelope set. So the algorithm runs in linear complexity. After obtaining the boundary points, we need to fill in the values for $\gamma(x)$. The difficulty is that some boundary points are not continuous, e.g., Figure \ref{fig:proof_case1} and Figure \ref{fig:proof_case4}. For Figure \ref{fig:proof_case1}, at the boundary point, we select the value given by the left parabola. And for Figure \ref{fig:proof_case4}, we select the value given by the right parabola. Note that Figure \ref{fig:proof_case2} and Figure \ref{fig:proof_case4} both belong to the second case where $s^* \in [r(k),r(k+1)]$. The difference is that in Figure \ref{fig:proof_case2}, we have $s^* = s$ since $u^{-1}(z) < s$, while in Figure \ref{fig:proof_case4}, we have $s^* = u^{-1}(z)$ since $u^{-1}(z) > s$.

\newpage

{\small
\bibliographystyle{ieee}
\bibliography{egbib}
}

\end{document}